\documentclass[10pt,twocolumn,letterpaper]{article}

\usepackage[pagenumbers]{cvpr} 

\usepackage{graphicx}
\usepackage{amsmath}
\usepackage{amssymb}
\usepackage{booktabs}
\usepackage{graphbox}
\usepackage{array}
\usepackage{multirow}
\usepackage{float}
\usepackage{xcolor}
\definecolor{darkpastelgreen}{rgb}{0.01, 0.75, 0.24}
\newsavebox\CBox
\def\textBF#1{\sbox\CBox{#1}\resizebox{\wd\CBox}{\ht\CBox}{\textbf{#1}}}

\newcommand{\F}[1]{\textBF{#1}}

\usepackage[pagebackref,breaklinks,colorlinks]{hyperref}

\usepackage[capitalize]{cleveref}
\crefname{section}{Sec.}{Secs.}
\Crefname{section}{Section}{Sections}
\Crefname{table}{Table}{Tables}
\crefname{table}{Tab.}{Tabs.}

\def\cvprPaperID{2907} 
\def\confName{CVPR}
\def\confYear{2023}

\begin{document}

\title{Sparse SPN: Depth Completion from Sparse Keypoints}
\author{{Yuqun Wu$^*$ 
\qquad
Jae Yong Lee$^*$ 
\qquad
Derek Hoiem} \\
University of Illinois at Urbana-Champaign\\
{\tt\small \{yuqunwu2, lee896, dhoiem\}@illinois.edu}
}
\maketitle
\def\thefootnote{*}\footnotetext{Indicates equal contribution}
\begin{abstract}
Our long term goal is to use image-based depth completion to quickly create 3D models from sparse point clouds, e.g. from SfM or SLAM.  Much progress has been made in depth completion. However, most current works assume well distributed samples of known depth, e.g. Lidar or random uniform sampling, and perform poorly on uneven samples, such as from keypoints, due to the large unsampled regions.  To address this problem, we extend CSPN with multiscale prediction and a dilated kernel, leading to much better completion of keypoint-sampled depth.  We also show that a model trained on NYUv2 creates surprisingly good point clouds on ETH3D by completing sparse SfM points.
    
\end{abstract}

\section{Introduction}
    
    Depth completion methods aim to produce a depth value for each pixel, given an RGB image and a sparse set of depth values. Sparse depths may come from depth sensors, such as Lidar, or feature-based reconstruction such as structure-from-motion (SfM) or simultaneous localization and mapping (SLAM). Completed depth maps are useful for many applications, including robotics grasping and navigation, novel view synthesis, photo editing, and augmented reality. Accurate and efficient completion is, therefore, an important area of research.
    
    Existing works~\cite{deep_depth_completion,deep_lidar,cspn,nlspn,cspn++,PENet,ACMNet}  focus on Lidar completion, where sparse depth is distributed roughly uniformly over the image. For experiments on indoor scenes, these methods simulate sparse depth by uniform random sampling from ground truth depth.  
    In both cases, each local neighborhood is likely to contain at least one sparse depth point, so existing works can be seen as RGB-guided interpolation of depth within a limited spatial extent.  
    
    \begin{figure}[t]
    \setlength{\belowcaptionskip}{-0.3cm}
    \centering
    \begin{tabular}{c@{\hskip 0.2em}c}
        \includegraphics[width=0.22\textwidth]{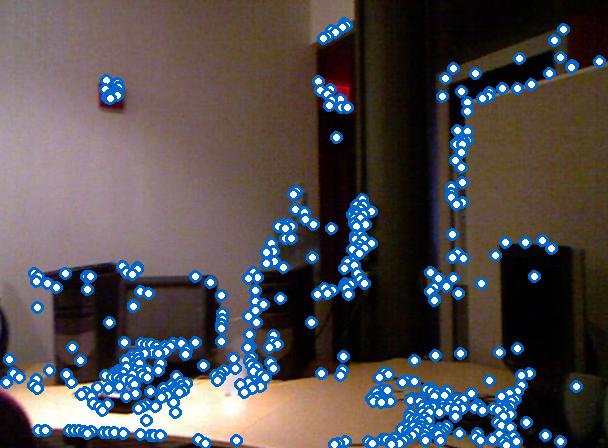}& 
        \includegraphics[width=0.22\textwidth]
        {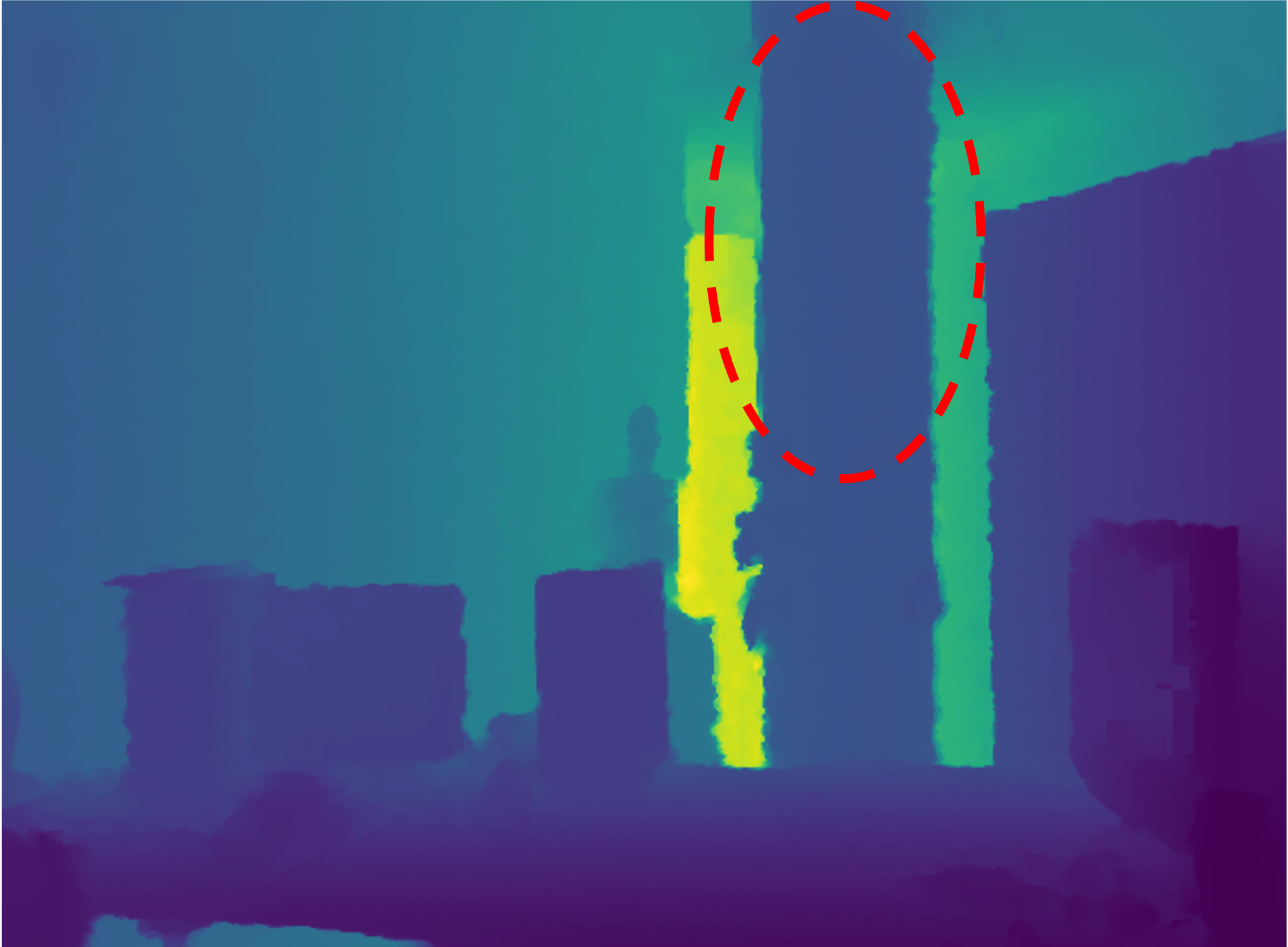} \\
        (a) RGB \& S. Depth & (b) Ground Truth \\
        \includegraphics[width=0.22\textwidth]{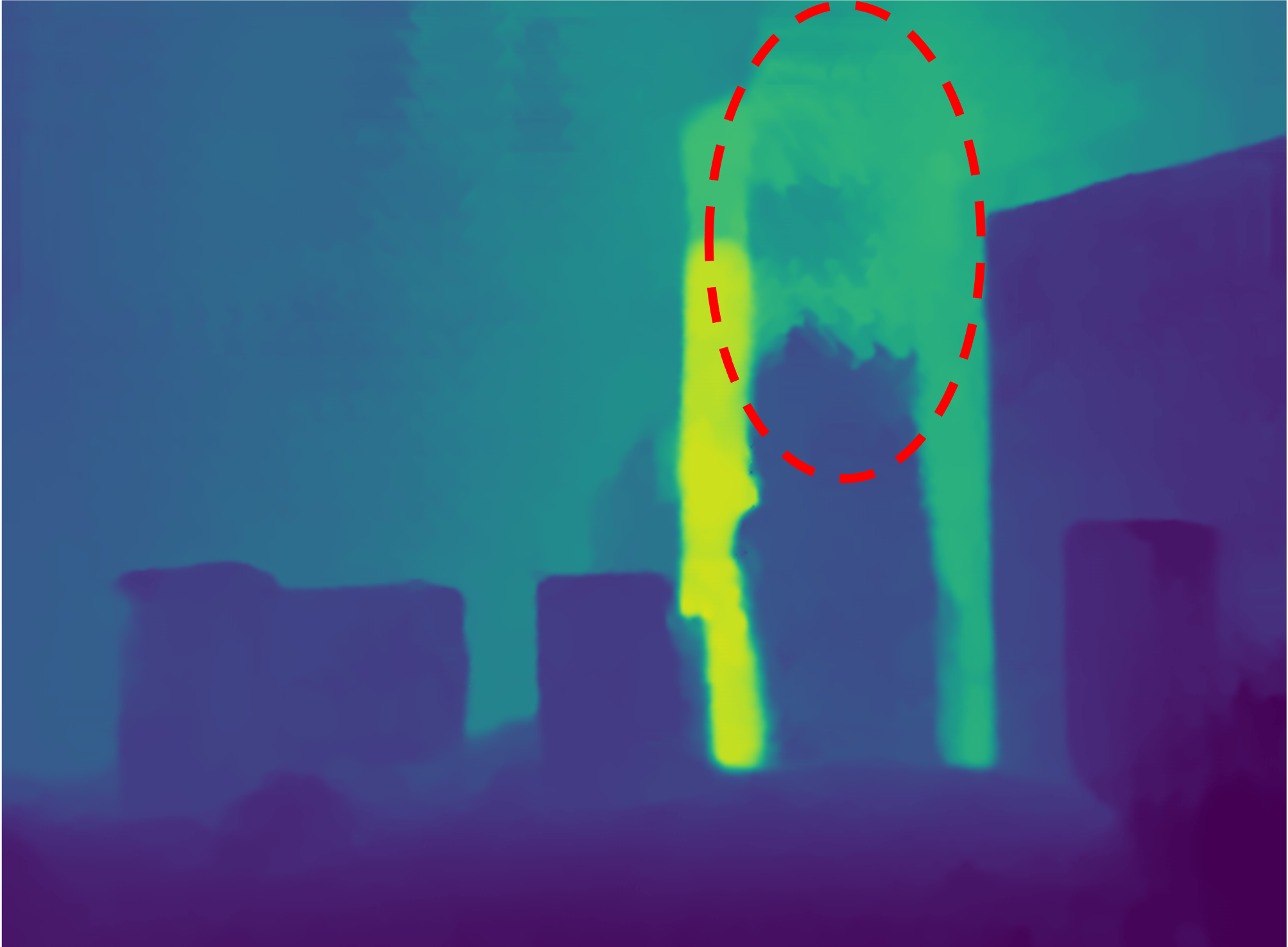} & 
        \includegraphics[width=0.22\textwidth]{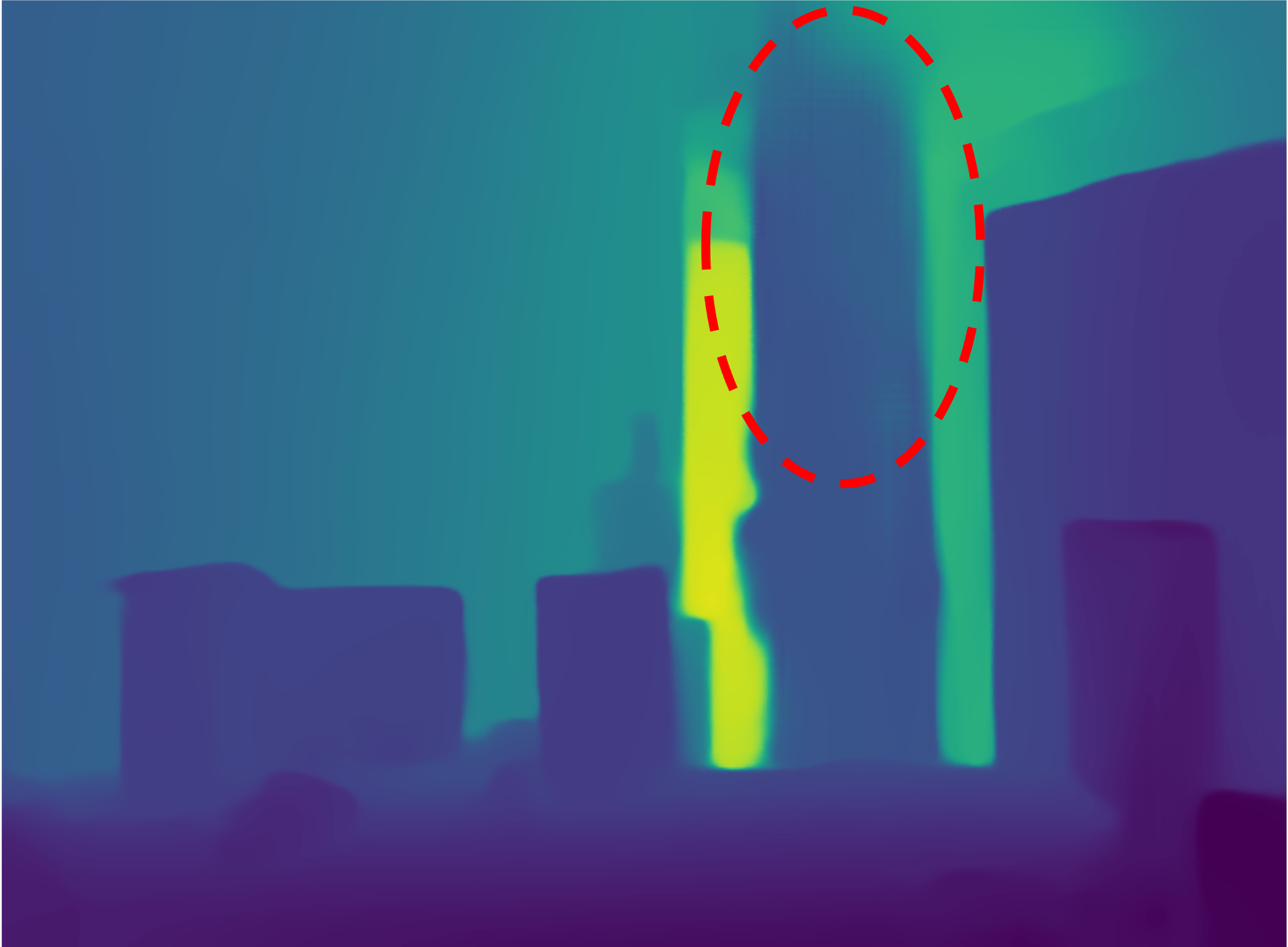} \\ 
         
         (c) CSPN~\cite{cspn}& (d) Ours \\
    \end{tabular}
    \caption{
    \label{fig:cpsn_failure_label} 
    \textbf{Depth completion from keypoint positions}. In the dashed red circle, CSPN~\cite{cspn} fails to complete depth due to its limited kernel receptive field, while our method recovers a complete and accurate depth map. 
    }
 \end{figure}

    Our work focuses on completing depth from sparse keypoint-based reconstruction, e.g. obtained by SfM or SLAM with SIFT~\cite{SIFT} or FAST~\cite{FAST} features. In such cases, sparse depth is typically densely distributed in textured regions, while unavailable in large untextured regions. We find that currently best-performing methods are unable to accurately propagate and extrapolate depth to these large unsampled regions (Fig.~\ref{fig:cpsn_failure_label}). We show that training on keypoint-sampled depth values significantly improves inference in keypoint-based depth while retaining similar performance in completion from uniformly sampled depth. Further, our proposed SSPN (Sparse Spatial Propagation Network) algorithm builds on Convolutional Spatial Propagation Network (CSPN)~\cite{cspn}, which learns to propagate depth values using a data-sensitive affinity matrix, by introducing coarse-to-fine prediction, dilating the kernel~\cite{dilation_koltun,Red_black_scheme}, and adding guidance from predicted surface normals. Together, these greatly increase the network's ability for long-range depth propagation.
    
    \begin{figure*}[t]  
    \centering 
    \includegraphics[width=0.98\textwidth]{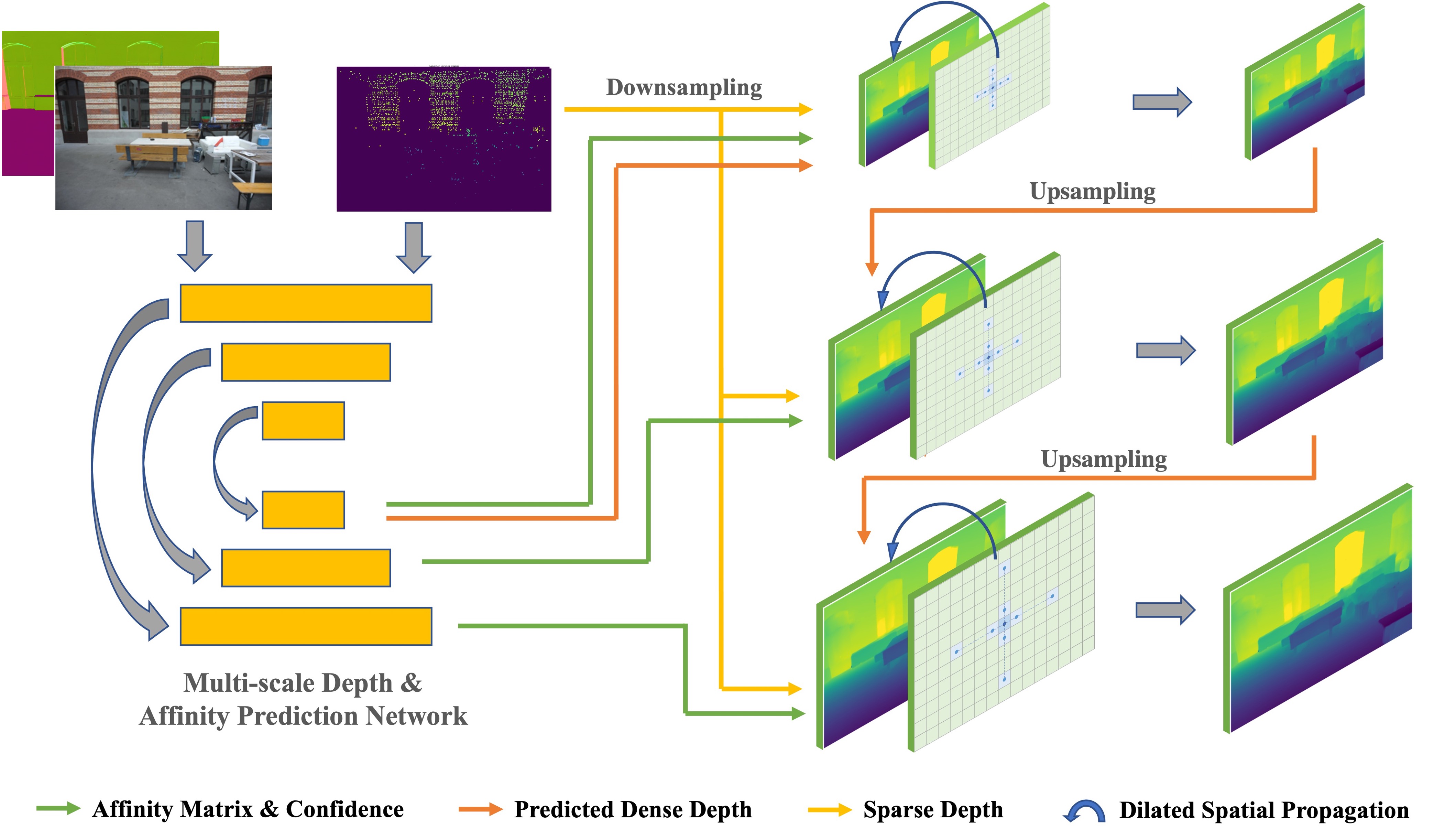}   
    \caption{\textbf{Structure of Sparse Spatial Propagation Network.}
    Our U-Net~\cite{Ronneberger2015UNetCN} shaped CNN network predicts the dense depth at the lowest resolution, and the affinity matrix and confidence map at each scale. 
    Then, series of dilated spatial propagations are applied at each scale to refine the depth maps in a coarse-to-fine manner. 
    Surface normal is predicted given RGB image using a pretrained model~\cite{kar20223d}. 
    Sparse depth is passed into each scale to better use the sparse input information.}
    \label{fig:Network}  
\end{figure*}     
    
    In experiments, we train our depth completion model on RGBD images from NYU v2 dataset~\cite{nyuv2} based on depth values from SIFT~\cite{SIFT} detected keypoint locations. Evaluating on NYU v2 test data, we show that our SSPN model outperforms CSPN~\cite{cspn} and NLSPN~\cite{nlspn}  methods in this setting.  
    We further demonstrate completion of sparse depth points by applying our model trained on NYU v2 to noisy sparse points obtained using SfM on the ETH3D dataset~\cite{eth3d}, again substantially outperforming other spatial propagation networks for depth completion.  
    We fuse the depth maps to obtain 3D point clouds and show that our method yields similar $F_1$ scores to the GIPUMA~\cite{Red_black_scheme} MVS algorithm, though with a very different accuracy-completion trade-off. Our ablation study indicates that dilation and coarse-to-fine modifications have little benefit on their own but jointly lead to large improvement, which made the discovery of this improvement non-obvious.  We also find that models trained on keypoint sampled depth are generally applicable, while those trained on uniformly sampled depth perform poorly in other sampling scenarios. Finally, we compare under the experimental setups described by previous works for visual SLAM completion on Azure Kinect Dataset~\cite{sartipi2020deep}, low resolution uniformly sampled depth completion on NYU v2 and Lidar depth completion on KITTI~\cite{kitti}, where our method is competitive with state-of-the-art.
  
    Our main \textbf{contributions} are to investigate and improve depth completion from unevenly sampled points (keypoints), including:
    \begin{itemize}
        \item \textbf{Propose dilated kernel, coarse-to-fine architecture, and surface normal guidance for SPN} that improve depth completion, in part by increasing the receptive field and providing geometry guidance for propagation
        \item \textbf{Evaluate depth completion from keypoint positions}, demonstrating that existing SPN-based depth completion methods are not suited to unevenly sampled points, that our proposed contributions yield significant improvements, and that efficient generation of dense point clouds from sparse SfM points is possible. 
    \end{itemize}

\section{Related Works}
\label{sec:related_work}

\noindent
\textbf{Approaches to Depth Completion: } 
\quad Depth completion aims to recover dense depth mapping given sparse depth values and an RGB image. 
Ma and Karaman~\cite{sparse_to_dense}, to our knowledge, are the first to apply deep learning to the problem.  They use a CNN encoder-decoder to generate depth maps from RGB and sparse depth values, showing large improvements compared to prediction from RGB only.  Application is also shown to Lidar completion and completion from sparse SLAM points, though the latter includes only one qualitative result. Experimenting on Lidar and simulated Lidar, 
Jaritz et al.~\cite{sparse_to_dense_with_semantics} show that separately encoding RGB and sparse depth is helpful, with late stage fusion before the decoder. Zhang and Funkhauser~\cite{deep_depth_completion} complete missing regions in sensor (e.g. Kinect) depth maps by predicting surface normals and boundaries and using them as constraints to solve for missing depth values.
Xiong et al.~\cite{sparse2dense_gnn} show the effectiveness of completion based on propagation using a Graph Neural Network (GNN). They also show that Poisson disc sampling of sparse depth enables better completion than uniformly random sampling, but this seems to assume that dense depth data is available at test time. 
Qiu et al.~\cite{deep_lidar} and Xu et al.~\cite{depth_normal} generate surface normals and confidence maps and use them together with the RGB and sparse depths to complete depth maps from Lidar sensors. In a refinement stage, Xu et al. also incorporate a diffusion block and guidance map, which is conceptually similar to the spatial propagation networks discussed in the next subsection.  Zhao et al.~\cite{ACMNet} introduce a co-attention guided graph propagation in encoder to enable pixels to capture useful observed contextual information more effectively, and uses symmetric gated fusion strategy to fuse the multi-modal contextual information efficiently, with application to Lidar and randomly sampled depth completion.   
Hu et al.~\cite{PENet} refine predictions from an encoder-decoder network with CSPN++ propagation~\cite{cspn++}.  
Huynh et al.~\cite{Huynh2021BoostingMD} explore depth completion with points from SfM in an RGBD indoor dataset by extracting multi-scale RGB and 3D sparse point features, but experiments do not include point cloud generation, and code is not available for comparison.

Motivated by depth completion from SLAM, Zhong et al.~\cite{zhong2019deep} propose a deep CCA method to recover missing depth values based on correlation between RGB features with known sparse depth values.  Their experiments include completion from uniform, high-texture, and feature-based sampling, demonstrating slight improvement over CSPN.  Another line of work~\cite{teixeira2020aerial,merrill2021robust,Zuo2021CodeVIOVO} provides efficient solution that targets dense depth estimation on light-weight, embedded systems.
These methods allows inference with small computation power, but at the cost of lower precision.
Wong et al. learn to complete depth in an unsupervised manner based on photometric consistency~\cite{wong2020unsupervisedICRA} and extend to incorporate adaptive regularization~\cite{Wong2021LearningTF_roboticsletters} and camera calibration~\cite{Wong2021UnsupervisedDCICCV}.
Sartipi et al.~\cite{sartipi2020deep} use planar geometry and gravity estimation to augment the sparse depth to perform depth completion network.  These approaches rely on creating planar prior, through triangulation~\cite{wong2020unsupervisedICRA} or surface normal estimation with masking~\cite{sartipi2020deep}. When compared on broadly used depth completion benchmarks, such as KITTI~\cite{kitti} and NYU v2~\cite{nyuv2}, these approaches, which are based on UNet-style architectures, underperform recent SPN-based approaches, though some may benefit from additional information such as IMU sensors or calibration.  Our approach extends the line of SPN approaches to achieve state-of-the-art on completion of non-uniformly sampled depth values, while learning from ground truth depth and not requiring additional information or priors beyond the RGB image and sparse depth.

\begin{figure}[t]
    \setlength{\belowcaptionskip}{-0.4cm}
    \centering
    \begin{tabular}{cccc}
        \includegraphics[width=0.09\textwidth]{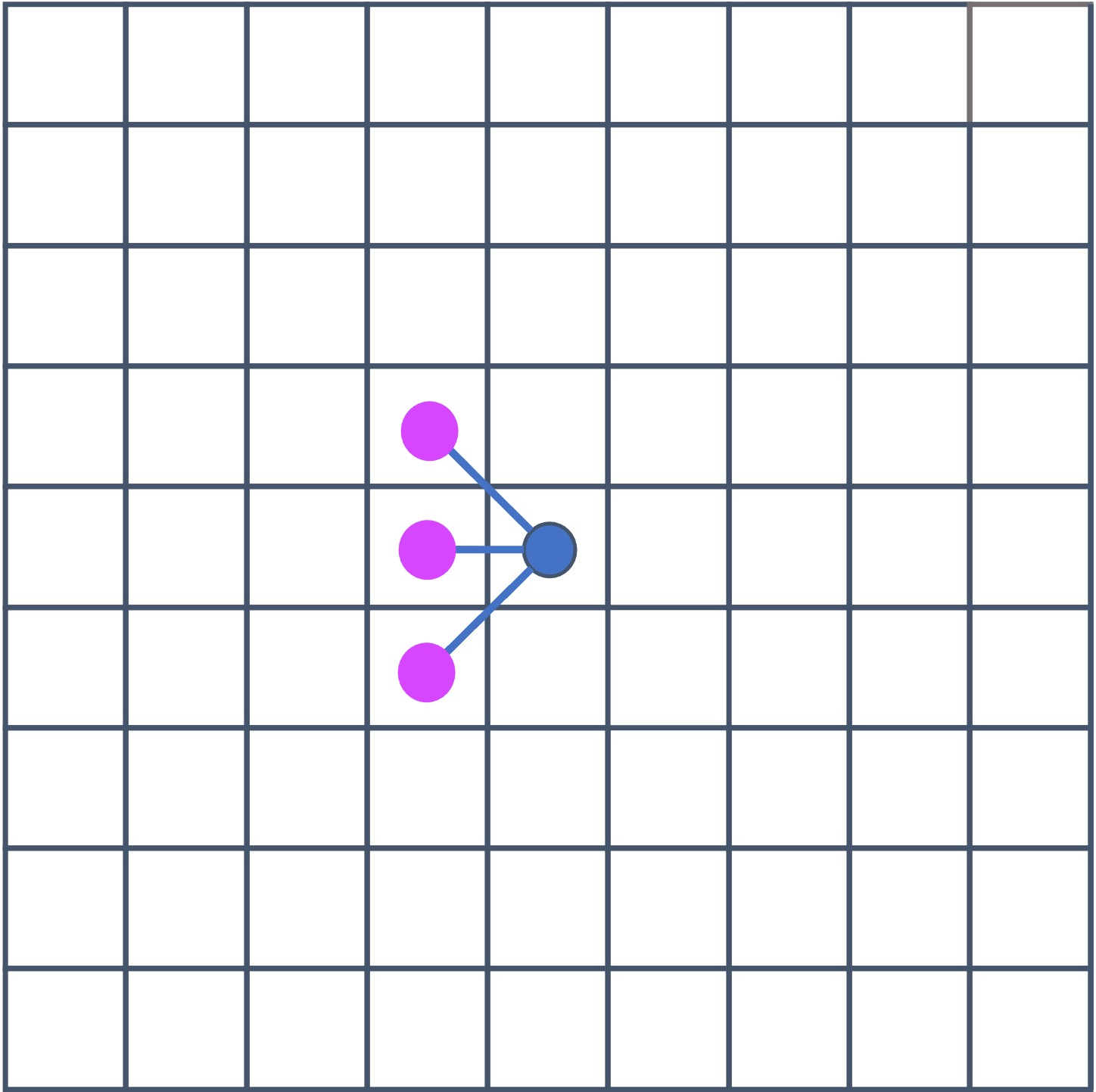} & 
        \includegraphics[width=0.09\textwidth]{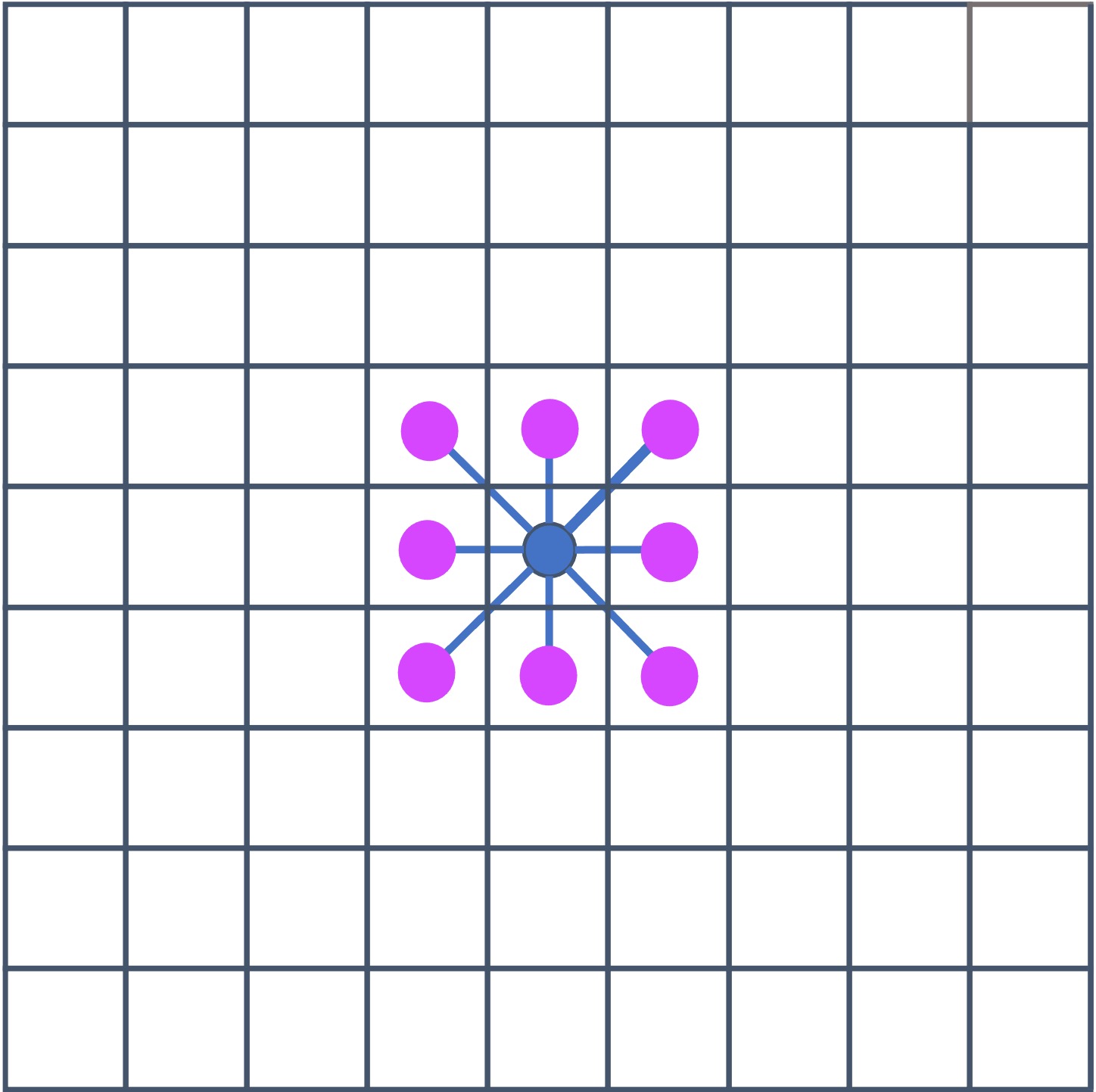} &
       \includegraphics[width=0.09\textwidth]{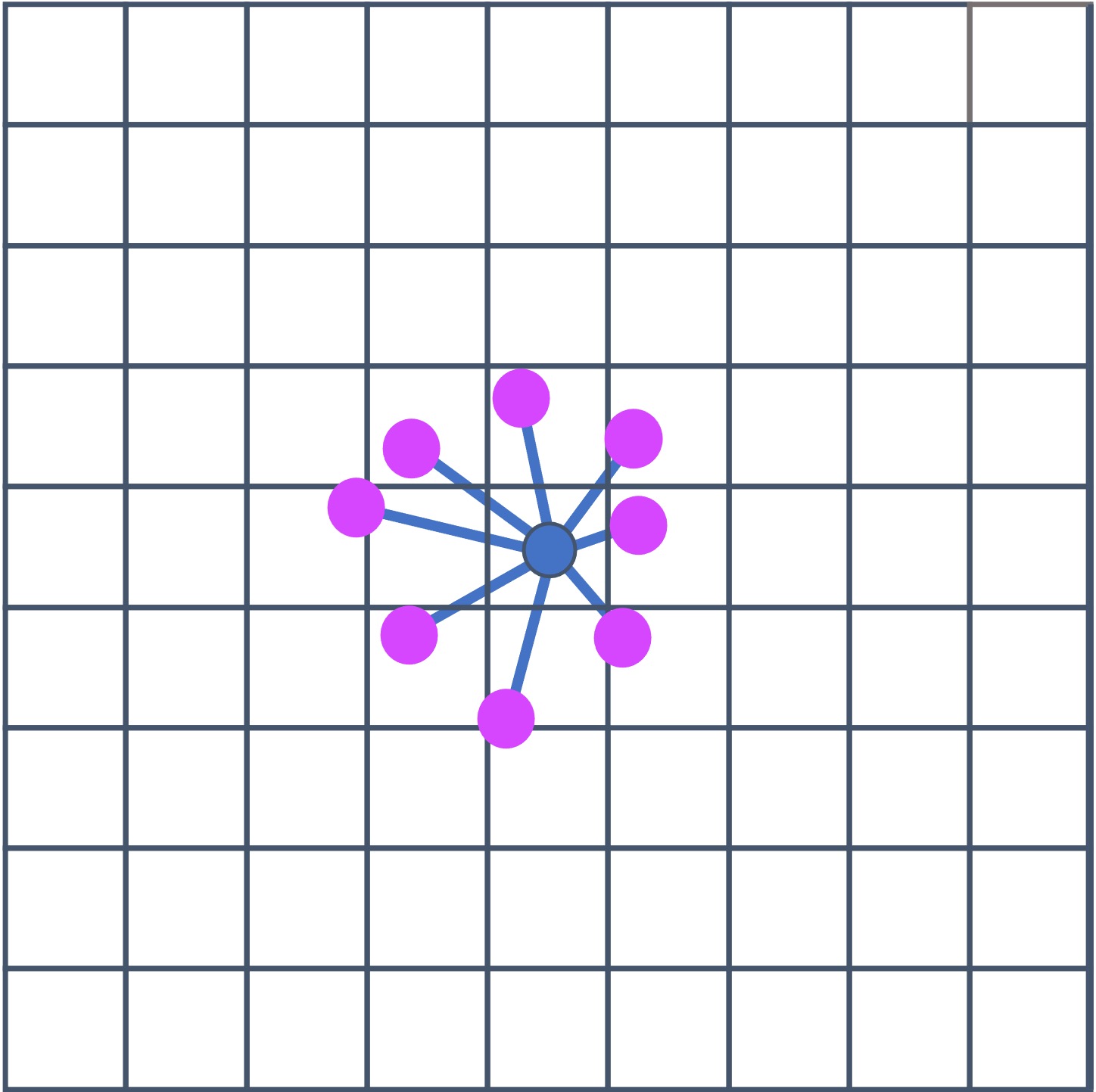} &
        \includegraphics[width=0.09\textwidth]{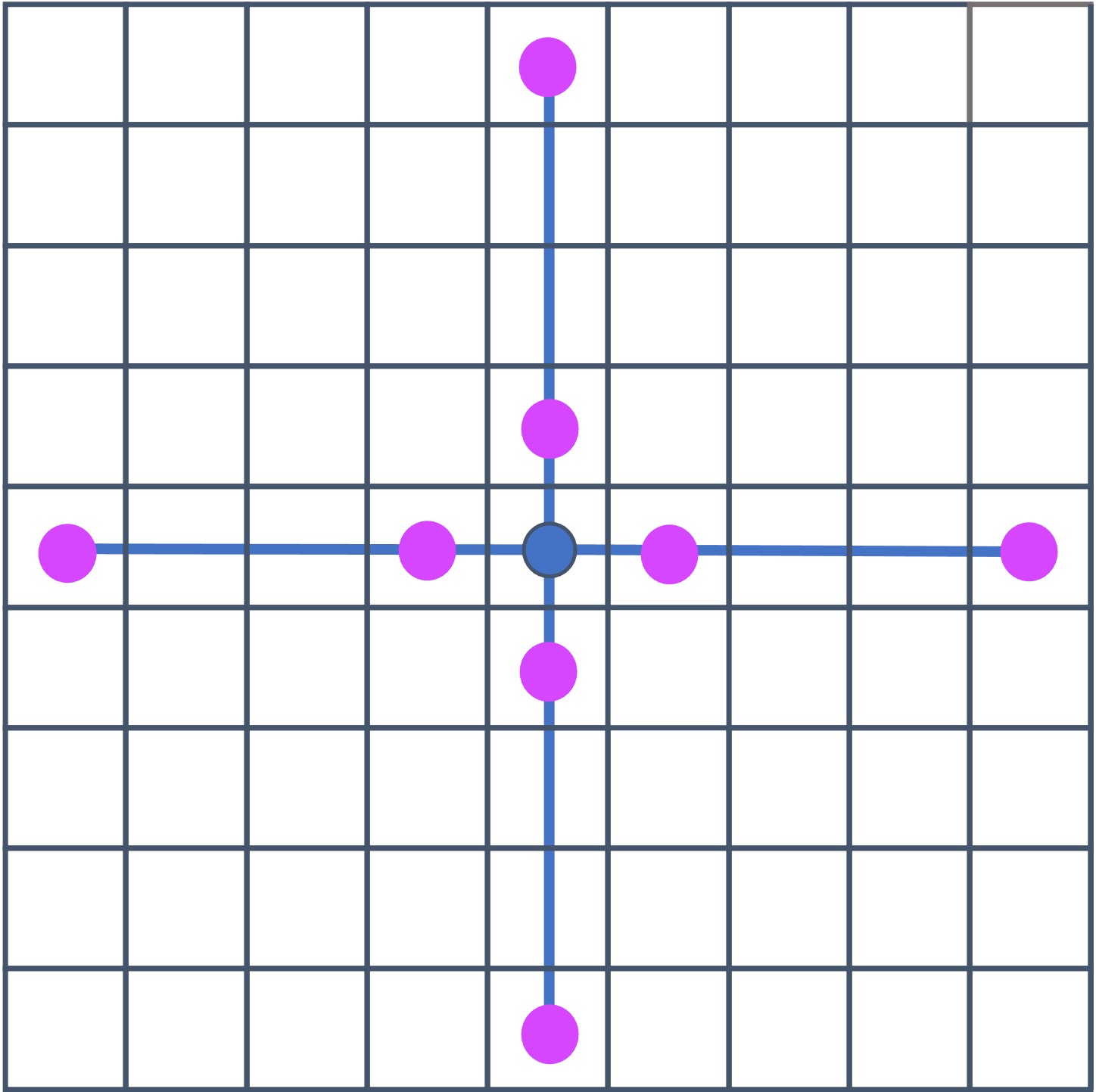} \\
           \footnotesize{(a) SPN\cite{spn}} &  \footnotesize{(b) CSPN\cite{cspn}} & \footnotesize{(c) NLSPN}\cite{nlspn} &  \footnotesize{(d) Ours}
    \end{tabular}
    \caption{\textbf{Different spatial propagation kernels}. Blue points are the center pixels, and purple points are the propagation candidates of each pipeline.
    }
    \label{fig:kernels} 
 \end{figure}

\noindent
\textbf{Spatial Propagation Networks: } 
\quad Liu et al.\cite{spn} propose spatial propagation networks (SPN).  These networks learn to predict affinity matrices that encode pairwise similarities, and use them to aggregate information through row-wise and column-wise propagation. 
To reduce the memory and time cost, SPN uses a three-way connection, which allows each pixel to receive information from three neighboring pixels in each propagation step. Since SPN cannot be fully parallelized, Cheng et al.~\cite{cspn} propose a convolutional version of spatial propagation called CSPN, in which each pixel's value is iteratively updated based on the neighborhood values from the previous iteration.   CSPN is applied to depth completion, with the affinity matrix predicted by the RGB image used to propagate sparse depth values. 
Cheng et al.\cite{cspn++} further improve the effectiveness and efficiency of CSPN by learning adaptive convolutional kernel sizes and the number of iterations for the propagation.  
Park et al.\cite{nlspn} use deformable convolution to enable adaptive, potentially non-local neighborhoods as the basis for propagation and also introduce an affinity matrix normalization scheme, leading to state-of-the-art performance for completion from NYU v2~\cite{nyuv2} random sampling. 
Lin et al.\cite{Lin2022DynamicSP} apply attention to affinity values of different distance, achieving state-of-the-art performance for KITTI~\cite{kitti} depth completion benchmark.

We extend CSPN by introducing coarse-to-fine estimation with dilated kernels, which we show outperforms NLSPN and other methods for completion from keypoint-sampled depth.

\section{Method}

We extend the Convolutional Spatial Propagation Network~\cite{cspn} with a coarse-to-fine scheme, dilated propagation kernel, and surface normal guidance.
Fig.~\ref{fig:Network} provides an overview of our method.
Given an RGB image, sparse depth, and (optionally) predicted surface normals map, our network predicts the initial depth map at the lowest resolution, and predicts an affinity matrix and confidence map at each scale.  Starting with the initial predicted depth, the depth is refined at each scale using spatial propagation with predicted affinity values and confidences and a dilated kernel and then upsampled. 

\begin{figure}[t]
    \centering
    \begin{tabular}{ccc}
        \includegraphics[width=0.135\textwidth]{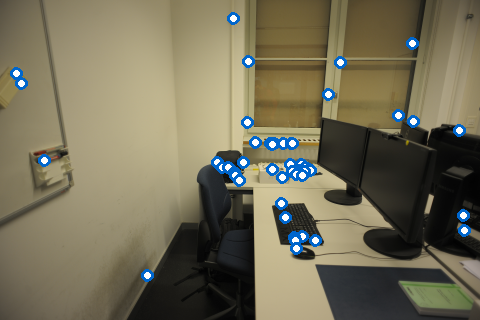} &
        \includegraphics[width=0.135\textwidth]{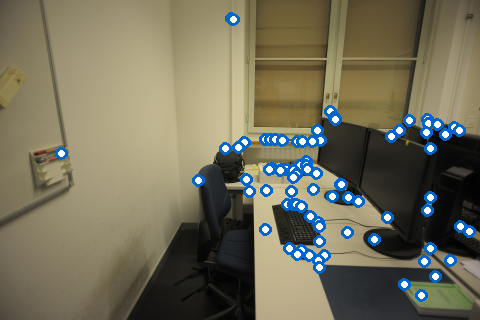} &
       \includegraphics[width=0.135\textwidth]{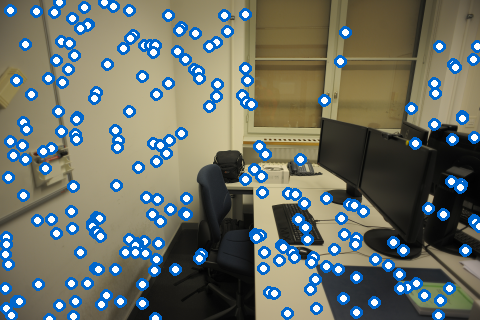} \\
           \footnotesize{(a) Sparse Points } & \footnotesize{(b) Keypoint } & \footnotesize{(c) Random} 
    \end{tabular}
    \vspace{-0.05in} 
    \caption{\textbf{Comparison of the sampling methods}. 
    The figure on the left presents the actual sparse points from SfM. 
    The right two figures show the depth locations from keypoint and random sampling.
    The keypoint samples are more similar to the sparse points, which are only available in textured areas (\eg keyboards, tissue boxes), while the random samples cover the whole image.
    }
    \label{fig:sample_comparison} 
\vspace{-0.1in}
 \end{figure}



\subsection{Convolutional Spatial Propagation Networks}
\label{sec:review_cspn}
Spatial propagation networks (SPN)~\cite{spn,cspn,nlspn} learn to predict affinity values that indicate pairwise similarity between each pixel and its neighbors. 
The pairwise similarities are used to iteratively propagate the known values to the nearby unknown values, such as propagating sparsely distributed depth values to the rest of the image~\cite{cspn,cspn++,nlspn}. 
The original SPN~\cite{spn} propagates along rows and columns.  Convolutional Spatial Propagation Network (CSPN)~\cite{cspn} updates all pixels simultaneously using convolution operations, based on values and affinites of each pixel's neighborhood.

For depth completion, CSPN takes image $I \rightarrow H \times W$ and the sparse depth map $S$ as input, and uses a U-Net~\cite{Ronneberger2015UNetCN} based CNN model to estimate an initial dense depth map $\mathcal{X} \rightarrow H \times W$ and the affinity map $\mathcal{W} \rightarrow H \times W \times C$, where $C$ denotes the number of neighbors. 
Formally, a single iteration of convolutional propagation can be defined as:

\begin{equation} \label{eq:propagation-without-conf}
\resizebox{.9\hsize}{!} 
    {$x_{i,j} = \begin{cases}
    w_{i,j}^c x_{i,j} + \sum_{(p,q) \in N(i,j)}{w_{i,j}^{p,q}x_{p,q}} & (i,j) \notin S \\
    s_{i,j} & (i,j) \in S \\
    \end{cases}$}
\end{equation}

where $x_{i, j}$ denotes the depth value at pixel coordinate ($i,j$), $N(i,j)$ denotes the set of neighboring pixel coordinates, $s_{i,j}$ denotes the value of input depth, $w_{i,j}^c$ denotes the weight of the center pixel, and $w_{i,j}^{p,q}$ denotes the affinity value between pixels $(i,j)$ and $(p,q)$. 
The weights $\textbf{w}_{i, j}$ are normalized to sum to 1. 
After fixed number of iterations, CSPN yields the refined dense depth map. 
The network is trained using the mean-squared-error loss between the refined depth map and ground truth depth map.

In more recent work~\cite{nlspn,cspn++}, confidences for each input depth value and predicted pixel are estimated.    
Given per-pixel confidence map $c_{i, j} \in \mathcal{C} \rightarrow H \times W$, per-input depth $(\forall (i,j) \in S), s_{i, j} \in \mathcal{S}$, and per-input depth confidence $(\forall (i,j) \in S), c^s_{i, j} \in \mathcal{C}{^s}$, spatial propagation using confidence can be defined as:
\begin{equation} \label{eq:propagation}
    x'_{i,j} =  w_{i,j}^c x_{i,j} + \sum_{(p,q) \in N(i,j)}{c_{p, q} w_{i,j}^{p,q}x_{p,q}}
\end{equation}
\begin{equation} \label{eq:sparse_propagation}
    x_{i,j} = \begin{cases}
    x'_{i,j} & (i,j) \notin S \\
    c^s_{i,j}s_{i,j} + (1 - c^s_{i,j}) x'_{i,j} & (i,j) \in S \\
    \end{cases}
\end{equation}

\begin{figure}[t]
    \centering
    \includegraphics[width=0.48\textwidth]{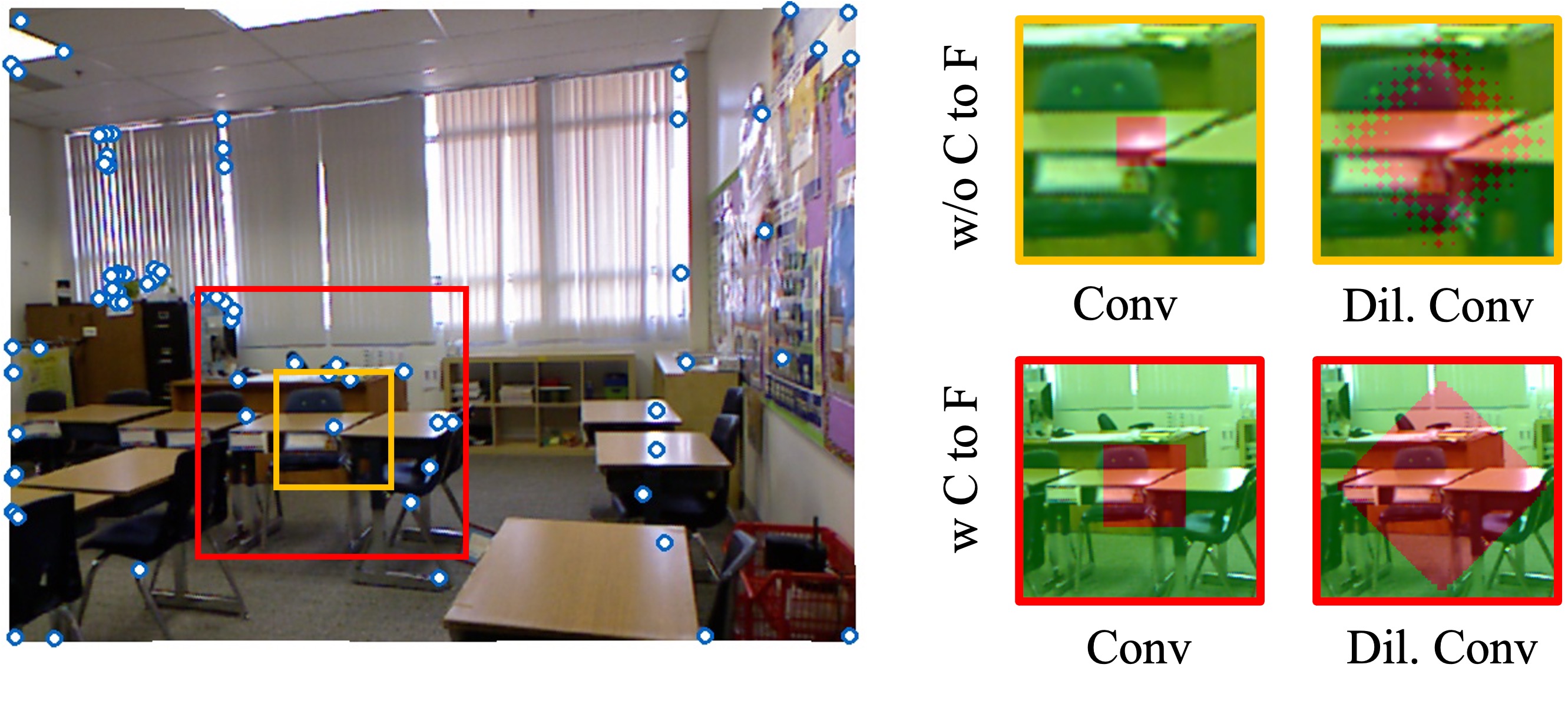} \\ 
    \vspace{-0.1in}
    \caption{ \textbf{Receptive fields of different propagation settings}. 
    We compare the receptive fields of the different propagation kernel settings. 
    The left most image shows the reference image, and the corresponding regions of the patch used in the images on the right. 
    The right four images show different settings and their receptive fields. ``Conv'' represents the kernel used by CSPN~\cite{cspn}, ``Dil. Conv'' represents the dilated propagation kernel described in Section~\ref{sec:sparse_spatial_propagation}. ``C to F'' indicates whether the propagation was done under coarse-to-fine scheme.
    For the visualization purpose, we use 8 total iterations without coarse-to-fine, and 2 iterations for 4 scales with coarse-to-fine.
    We show that dilated spatial propagation has larger reachability with and without coarse-to-fine, and that using coarse-to-fine expands the reachability much further.
    }
    \label{fig:receptive_field_figure} 
\vspace{-0.1in}
 \end{figure}

\begin{figure*}[t]
    \centering
    \begin{tabular}{ccccc}
        \includegraphics[width=0.17\textwidth]{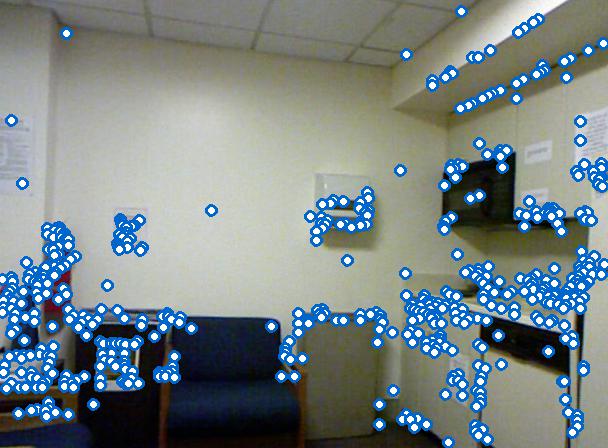}&
        \includegraphics[width=0.17\textwidth]{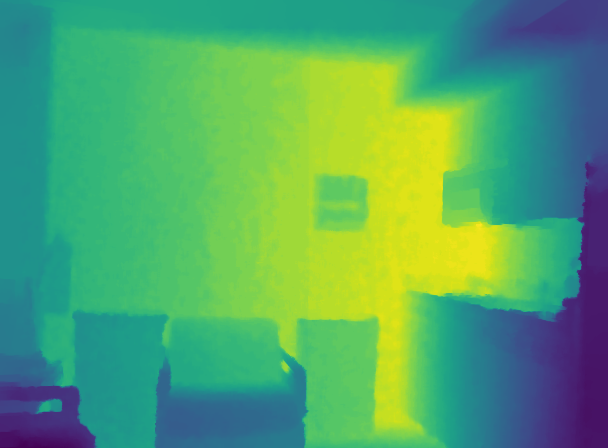}&
        \includegraphics[width=0.17\textwidth]{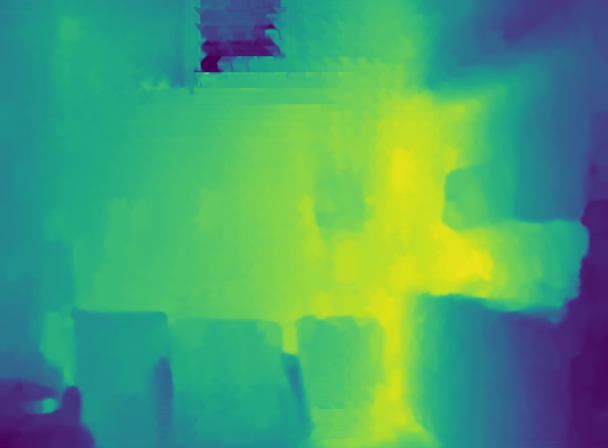}&
        \includegraphics[width=0.17\textwidth]{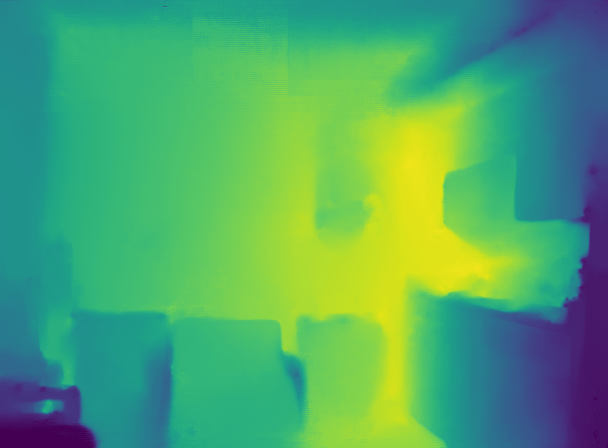}&
        \includegraphics[width=0.17\textwidth]{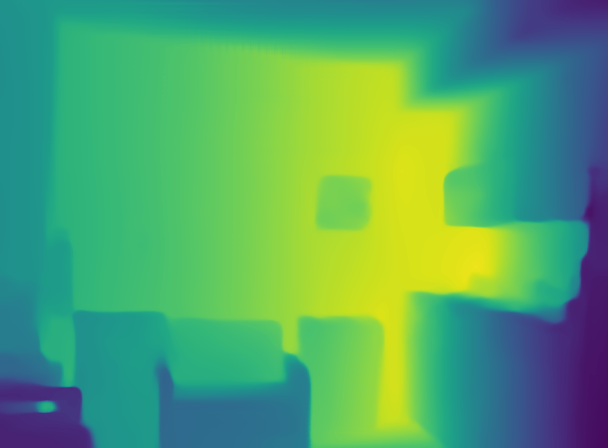} \\ 
        \includegraphics[width=0.17\textwidth]{images/new_s_depth/nyu_13_s_depth.jpg}&
        \includegraphics[width=0.17\textwidth]{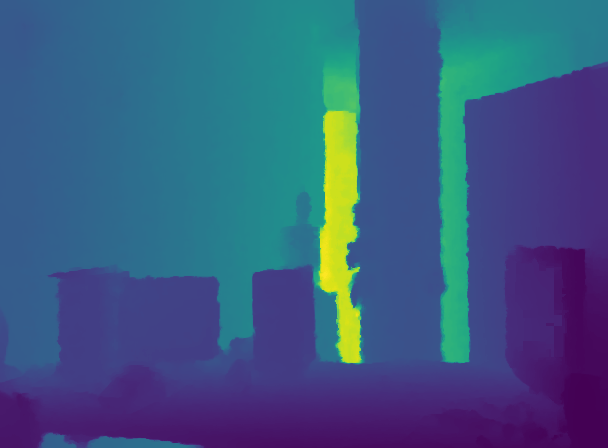}&
        \includegraphics[width=0.17\textwidth]{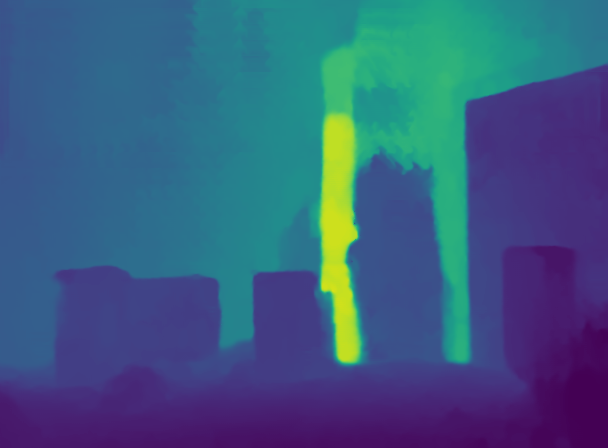}&
        \includegraphics[width=0.17\textwidth]{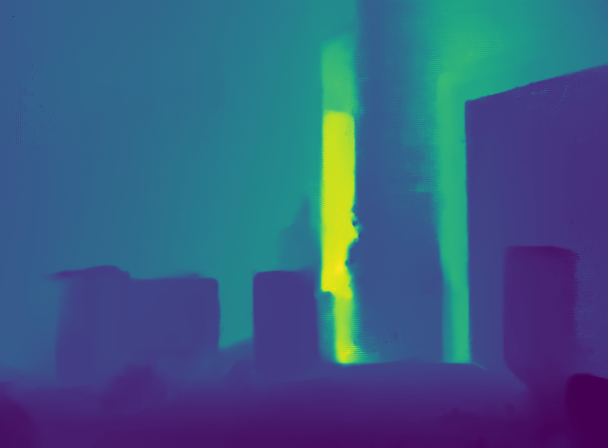}&
        \includegraphics[width=0.17\textwidth]{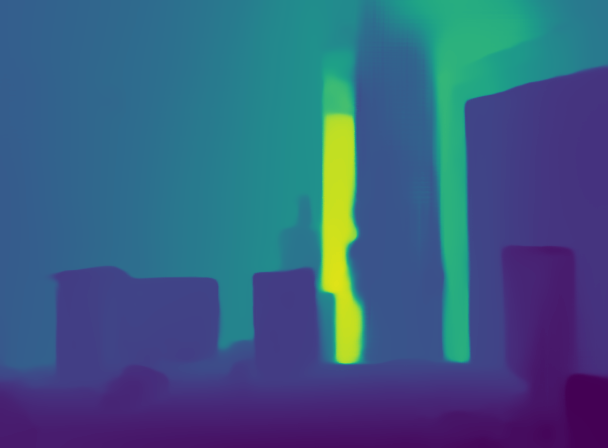} \\ 
        RGB \& Sparse input & Ground Truth &  CSPN\cite{cspn} & NLSPN\cite{nlspn} & Ours
    \end{tabular}
    \caption{\textbf{Qualitative comparison on NYUv2 with 800 keypoints input}. From the left, color images \& sparse input, ground truth, prediction from CSPN~\cite{cspn}, prediction from NLSPN~\cite{nlspn}, prediction of our method. Our method manages to recover areas that do not have any depth input (\eg walls, pillars), where all baselines fail to have reliable prediction.
    }
    \label{fig:nyuv2_qualitative_figure} 
 \end{figure*}
 \vspace{-0.1in}

\subsection{Keypoint Sampling}
\label{sec:keypoint_sampling}
When training and testing based on RGBD images, many methods employ uniformly random sampling (each pixel is equally likely) to obtain sparse depth values, which results in known values that are scattered over the entire image. We investigate keypoint-based sampling, in which the sparse depth values are sampled at the positions of detected SIFT~\cite{SIFT} keypoints. This simulates access to sparse depth values that may be available from keypoint triangulation, such as from SfM or SLAM.  Our experiments show that models trained on RGBD images using keypoint-based sampling can be effectively applied complete depth images using the output of an SfM algorithm, even on significantly different image sets from training.  SIFT detection is run on grayscale images, and the number of samples is capped.
As shown in Fig.~\ref{fig:sample_comparison}, the keypoints are densely distributed in highly textured regions while absent in other regions, which makes depth completion more difficult and particularly requires propagation over larger ranges, which motivates our Sparse Spatial Propagation Network (SSPN).

\subsection{Sparse Spatial Propagation Network}
\label{sec:sparse_spatial_propagation}
To expand the receptive field of propagation kernel, we propose two changes to the confidence-weighted CSPN architecture that was described in Sec.~\ref{sec:review_cspn}.   
First, we use coarse-to-fine scheme to extend the reach of the kernel by operating at low resolution in early iterations and refining into higher resolution.  
Next, we adopt the idea of diffusion based PatchMatch kernel~\cite{Red_black_scheme} using dilated convolution, which provides a larger receptive field without increasing the number of iterations or computational cost. 
To provide a geometry guidance for the network, our network takes additional  surface normal as input, which is estimated given color images with pretrained Omnidata network~\cite{kar20223d}.

Fig.~\ref{fig:Network} provides an overview of the architecture. 
Similar to CSPN~\cite{cspn}, we use U-Net~\cite{Ronneberger2015UNetCN} shaped model that receives an RGB image, surface normal image, and the corresponding sparse depth map and outputs an initial dense depth map at the coarsest scale and affinity and confidence maps at each scale. Within each scale, spatial propagation is iteratively performed using the dilated kernel shown in Fig.~\ref{fig:kernels}(d).
After a fixed number of iterations, the current dense depth map is bilinearly upsampled, and spatial propagation continues at higher resolution.   
The kernel has a smaller dilation offset at coarser scales (see Sec.~\ref{sec:implementation_details} for details).
Fig.~\ref{fig:receptive_field_figure} visualizes the impact of each modification.
By combining the coarse-to-fine architecture and the dilated kernel, the receptive field of the spatial propagation of our method (using 8 iterations over 4 scales), is increased by 93 times compared to the original CSPN method (using 24 iterations at the finest scale).



\subsection{Loss Function}
\label{sec:loss}
During the training, we apply a multi-scale $\ell_1$ loss between our prediction depth and ground truth:
\begin{equation}
    Loss = \sum_{s \in scale}{w_s * | d_{s}^{pred} - d_{s}^{gt} |},
\end{equation}
\label{eq:loss}
where $s$ is the scale; 
$d_{s}^{pred}$ is the predicted depth after propagation refinement in scale $s$; 
$d_{s}^{gt}$ is the ground truth resized to the corresponding scale $s$;
$w_s$ is the weight at scale $s$, which is $1$ for lowest resolution, and increases by 1 for each finer scale.
A smaller weight is used in low resolutions because the ground truth depth is less precise in low resolutions, due to downsampling. The use of multi-scale loss helps the network converge faster during training.

\begin{table}[t]
\centering
\resizebox{83.5mm}{!}{
\begin{tabular}{cc|cccccc}
\toprule
Train & Test & Method  & RMSE & REL & $\delta_{1.02}$ & $\delta_{1.05}$ & $\delta_{1.10}$  \\ 
\midrule
\multirow{3}{*}{Key} &
\multirow{3}{*}{Key} & 
   CSPN~\cite{cspn}     & 0.220         & 0.043         & 55.3          & 78.3          & 88.9          \\
&& NLSPN~\cite{nlspn}   & 0.179         & 0.036         & 60.3          & 80.8          & 90.9          \\
&& Ours                 & \textbf{0.147}& \textbf{0.026}& \textbf{70.2} & \textbf{87.6} & \textbf{94.4}
\\ 
\midrule
\multirow{3}{*}{Rnd} & \multirow{3}{*}{Rnd} & 
   CSPN~\cite{cspn}     & 0.105	        & 0.014         & 86.3          & 95.0          & 97.8          \\
&& NLSPN~\cite{nlspn}   & \textbf{0.098}& \textbf{0.012}& \textbf{88.6} & \textbf{95.6} & \textbf{98.0} \\
&& Ours                 & 0.102         & 0.013         & 86.5          & 94.6          & 97.6          \\ 
\midrule

\multirow{3}{*}{Key} & \multirow{3}{*}{Rnd}  & 
   CSPN~\cite{cspn}     & 0.131         & 0.020         & 71.1          & 92.3          & 97.1          \\ 
&& NLSPN~\cite{nlspn}   & 0.257         & 0.018         & 84.8          & 93.9          & 97.0          \\ 
&& Ours                 & \textbf{0.103}& \textbf{0.013}& \textbf{87.8} & \textbf{95.3} & \textbf{97.8} \\ 
\midrule
\multirow{3}{*}{Rnd} & \multirow{3}{*}{Key} &
   CSPN~\cite{cspn}     & 1.176         & 0.286         & 46.7          & 62.5          & 71.6          \\ 
&& NLSPN~\cite{nlspn}   & \textbf{0.330}& 0.083& \textbf{53.0} & \textbf{70.5} & \textbf{80.9} \\ 
&& Ours              & 0.406         & \textbf{0.078}         & 51.3          & 69.0          & 78.9          \\

\bottomrule
\end{tabular}
}
\caption{\textbf{Depth Completion Results on NYUv2 Test set.} 
    We evaluate CSPN~\cite{cspn}, NLSPN~\cite{nlspn} and our model on the NYUv2 test set by using different permutations of sampling strategy used in the training and the test time. 
    Column \emph{Train} and \emph{Test} indicates the different sampling schemes used for training and testing the model respectively
    \emph{Key} refers to the keypoint sampling and \emph{Rnd} refers to the uniform random sampling. 
    Numbers in \textbf{bold} are the best results for each settings. We show that our method outperforms both baselines when trained on the keypoint samples. 
}
\label{tab:nyuv2_whole_table}
\end{table}

\begin{figure*}[t]
    \centering
    \begin{tabular}{ccccc}
        \includegraphics[width=0.18\textwidth]{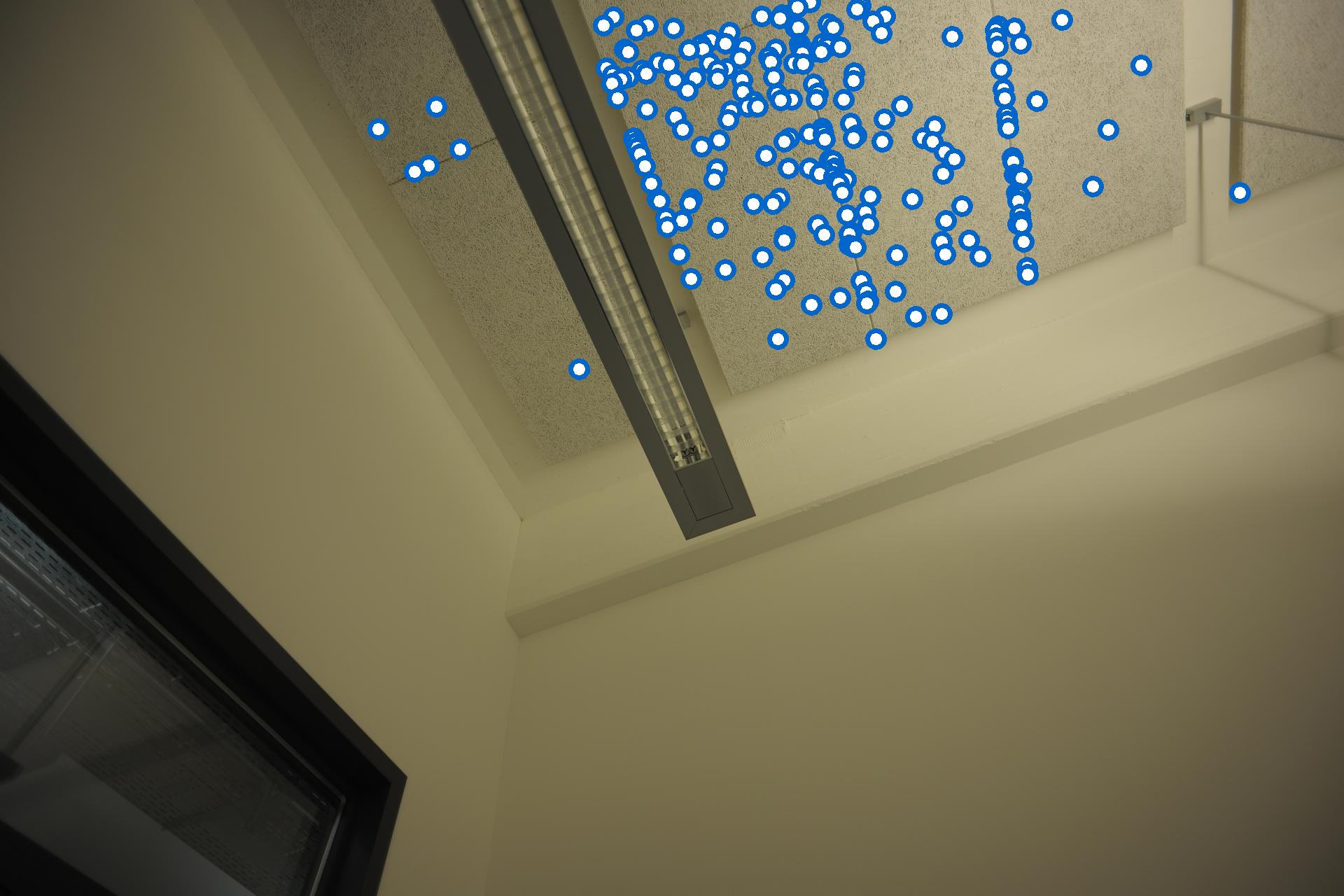}&
        \includegraphics[width=0.18\textwidth]{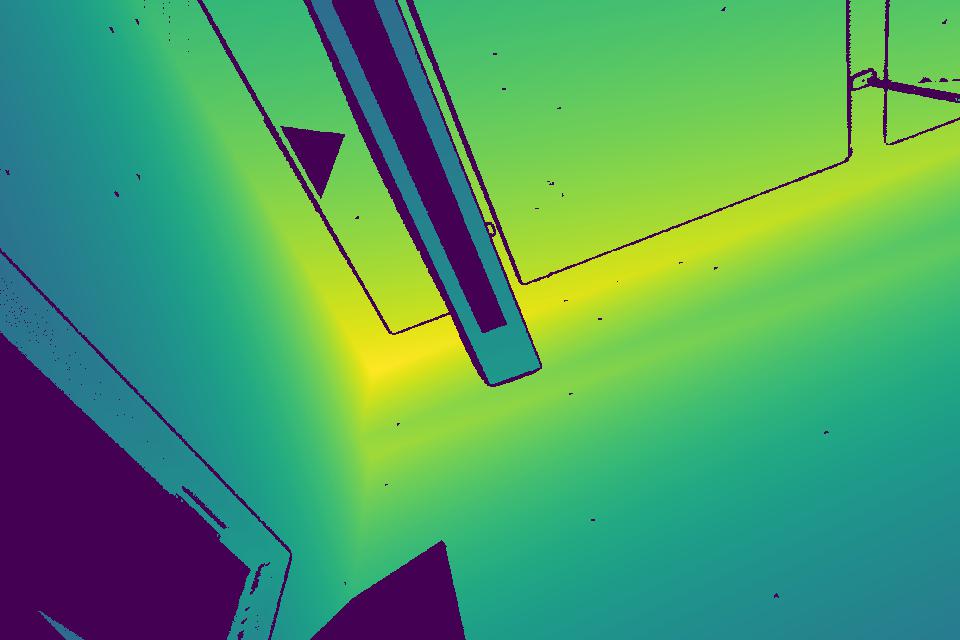}&
        \includegraphics[width=0.18\textwidth]{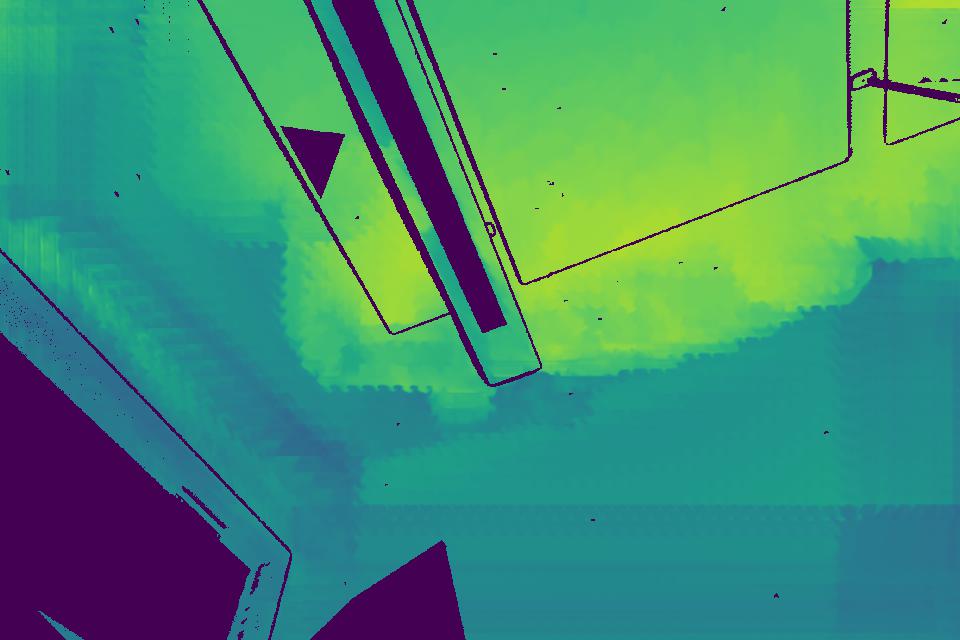}&
        \includegraphics[width=0.18\textwidth]{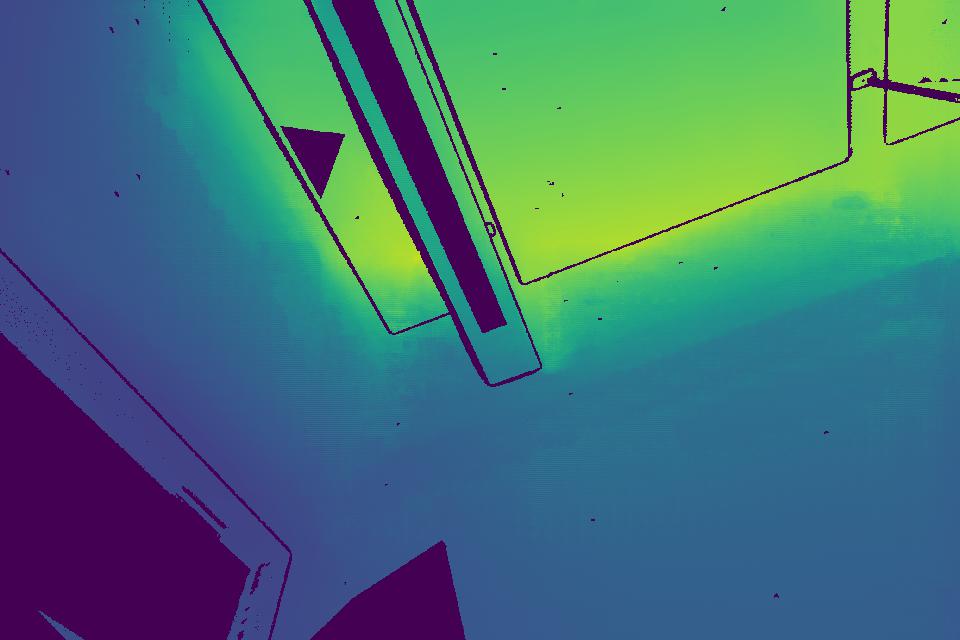}&
        \includegraphics[width=0.18\textwidth]{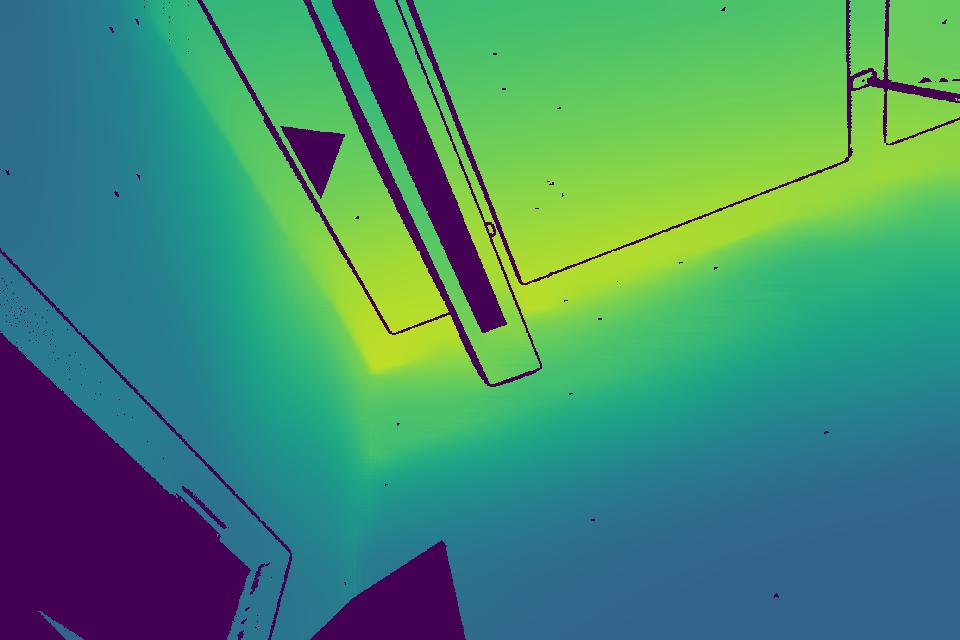}  \\
        \includegraphics[width=0.18\textwidth]{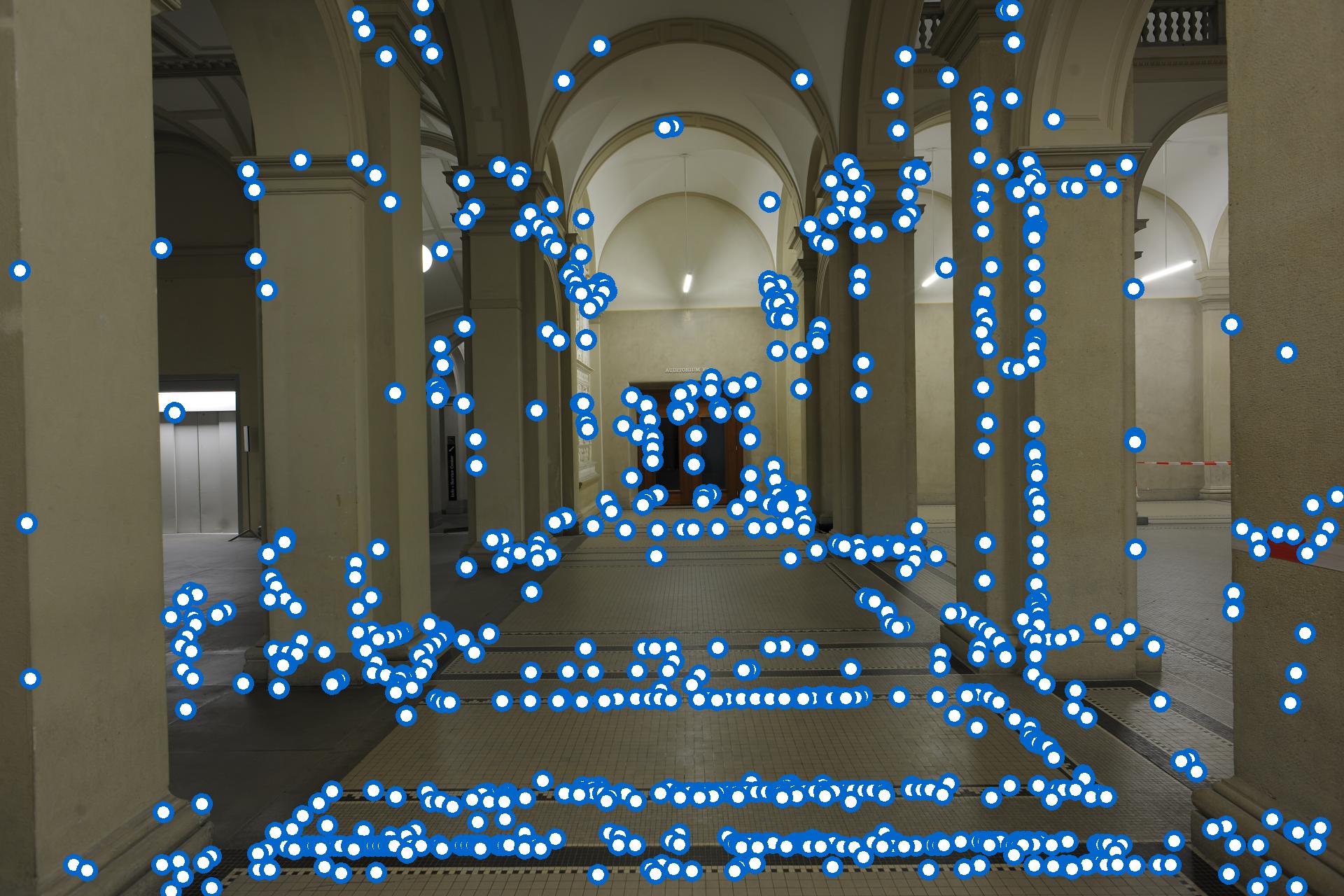}&
        \includegraphics[width=0.18\textwidth]{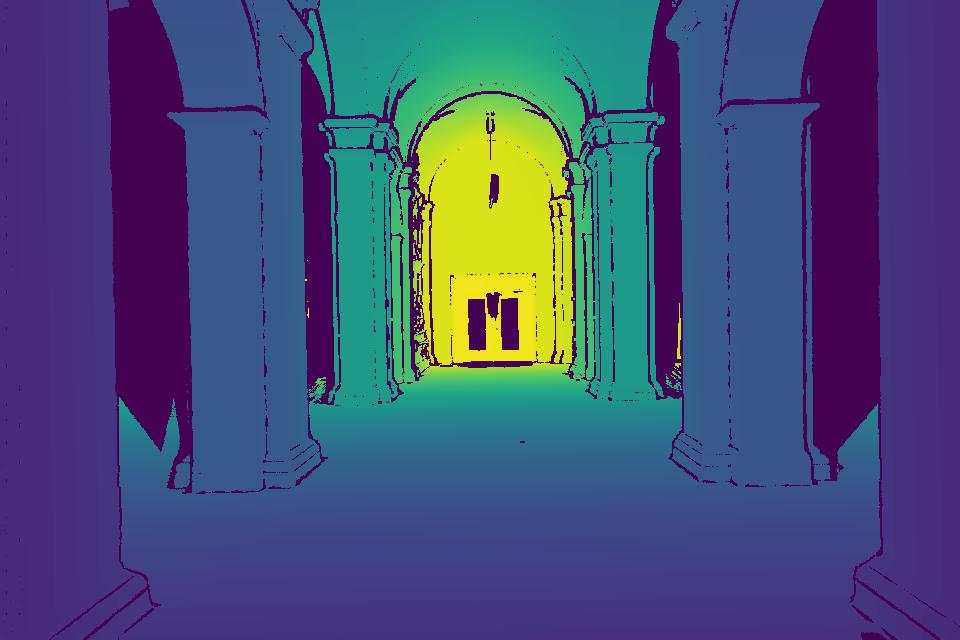}&
        \includegraphics[width=0.18\textwidth]{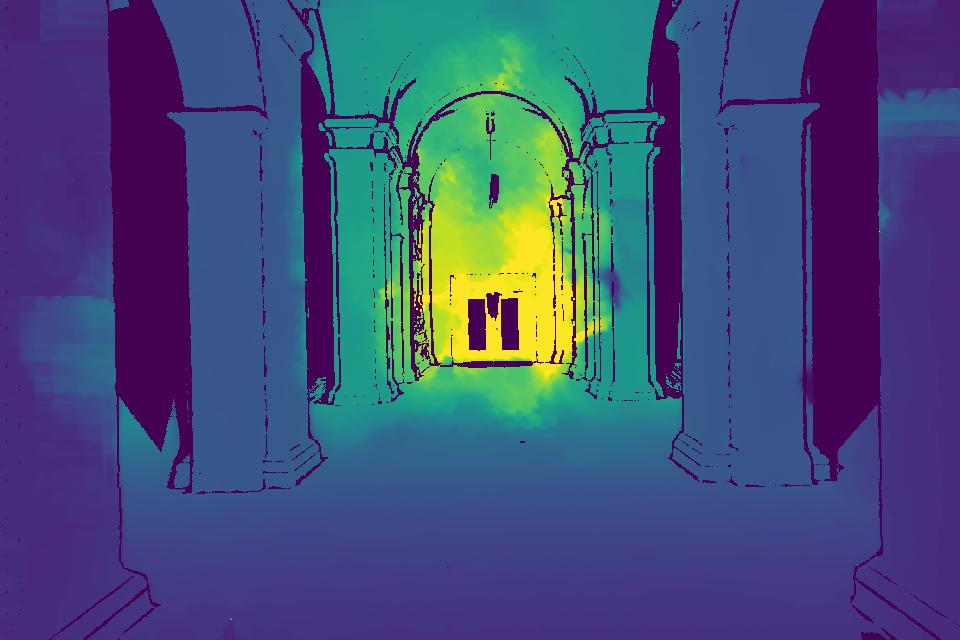}&
        \includegraphics[width=0.18\textwidth]{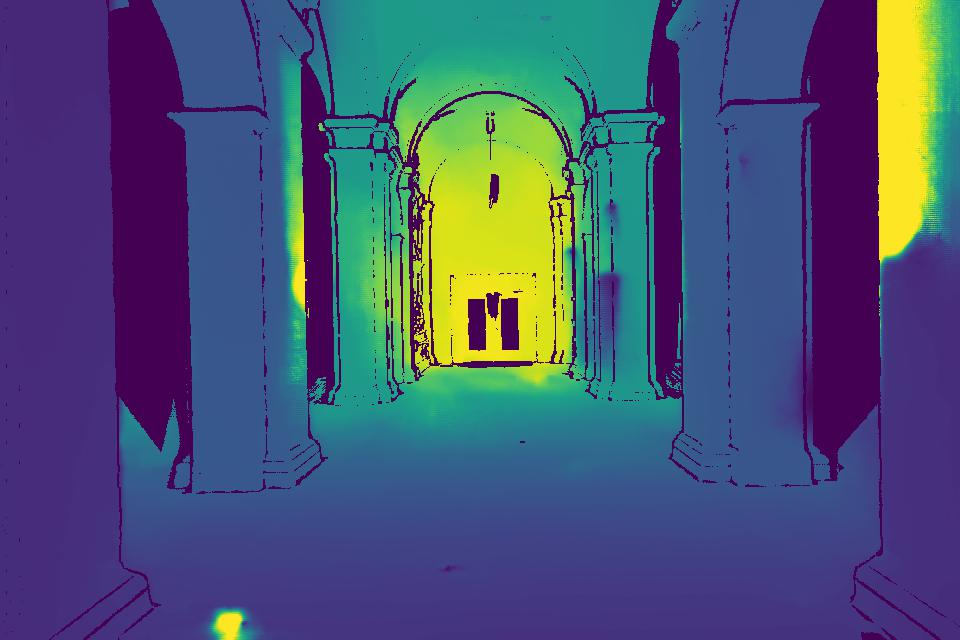}&
        \includegraphics[width=0.18\textwidth]{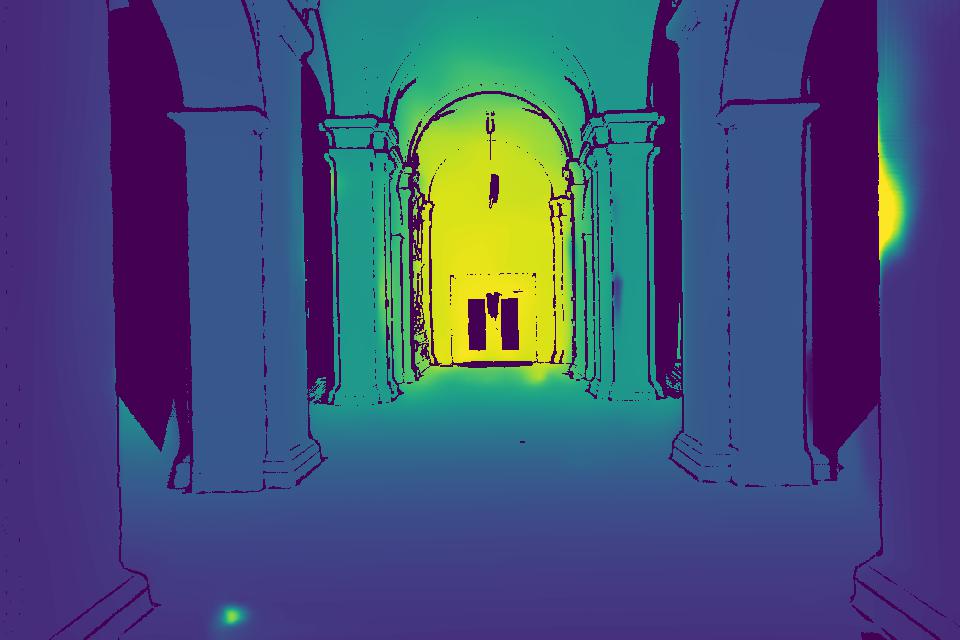}  \\
        RGB \& Sparse input & Ground Truth &  CSPN\cite{cspn} & NLSPN\cite{nlspn} & Ours
    \end{tabular}
    \caption{\textbf{Qualitative comparison on ETH3D}. 
    The number of input sparse points is low and some inputs are outliers, which leads to blocky artifacts in CSPN~\cite{cspn} and sometimes large error regions in NLSPN~\cite{nlspn}. 
    However, our method is typically more robust to outliers and generally yields more accurate depth maps.  
    }
    \label{fig:eth3d_depth_figure}
\vspace{-0.1in}
 \end{figure*}

\subsection{Implementation Details}
\label{sec:implementation_details}
We use an encoder-decoder network for all experiments, with ResNet34~\cite{Resnet} as the encoder, and a decoder that contains two convolution layers in each scale.
For all SPN methods without coarse-to-fine, 24 iterations of propagation are performed.
For coarse-to-fine networks, 4 scales and 8 iterations per scale are used. 
We use a dilation size of 2 for dilated propagation in lowest scale, and increase by 1 for each finer scale.
When sampling keypoint depth, we use the SIFT~\cite{opencv} detector of OpenCV~\cite{opencv}.

During the training, we use Adam~\cite{Adam} optimization with the initial learning rate as 0.001 and decay as 0.85 each epoch step for all experiments. 
The training batch sizes are 6 for NYUv2~\cite{nyuv2}, and 12 for KITTI~\cite{kitti}.
We train 30 epochs for all experiments.
We implement our code using PyTorch~\cite{pytorch}, and use 1 NVIDIA A40 on KITTI~\cite{kitti} and 2 NVIDIA TITAN X (Pascal) on remaining for training and testing.

\section{Experiments}

Our main experimental questions:
\begin{itemize}
    \item Are state-of-the-art depth completion methods effective when input depths are unevenly distributed? (see experiments on NYU v2: Sec.~\ref{sec:depth_completion_keypoints_nyu}, Table~\ref{tab:nyuv2_whole_table}, Fig.~\ref{fig:nyuv2_qualitative_figure})
    \item Do our proposed modifications to increase receptive field of propagation lead to significant improvement? (additionally, see ablations: Sec.~\ref{sec:ablation_studies_experiments}, Table~\ref{tab:ablation_study_table}) 
    \item When trained on keypoint-based sampling, can depth completion be used as a way to quickly generate dense point clouds from sparse SfM inputs? (see experiments on ETH3D: Sec.~\ref{sec:depth_completion_keypoints_eth3d}, Table~\ref{tab:eth3d_pc}, Fig.~\ref{fig:eth3d_pc})
\end{itemize}
For completeness, we also compare to recent approaches using experimental setups from the literature in Sec.~\ref{sec:Previous_evaluation_experiment}. 

\begin{table*}
\vspace{-0.1in}
\centering

\resizebox{\textwidth}{!}{
\begin{tabular}{llccccc@{\hskip 0.10in}ccccc@{\hskip 0.10in}ccccc@{\hskip 0.10in}ccccc@{\hskip 0.10in}ccccc@{\hskip 0.10in}ccccc}
\toprule
 &  & \multicolumn{3}{c}{\textbf{2cm: Completeness / Accuracy / F1}} & \multicolumn{3}{c}{\textbf{5cm: Completeness / Accuracy / F1}} \\
\textbf{Method} & \textbf{Resolution}  &  
\multicolumn{1}{c}{\textbf{Indoor}} & 
\multicolumn{1}{c}{\textbf{Outdoor}} & 
\multicolumn{1}{c}{\textbf{Combined}} &
\multicolumn{1}{c}{\textbf{Indoor}} & 
\multicolumn{1}{c}{\textbf{Outdoor}} & 
\multicolumn{1}{c}{\textbf{Combined}}\\
\midrule
Gipuma~\cite{Red_black_scheme}  & 2000x1332 & 
24.6   / {89.3} / {35.8} & 
25.3   / {83.2} / {37.1} & 
24.9   / {86.5} / {36.4} & 
34.0   / {96.2} / 47.1     & 
36.7   / {95.5} / {51.7} &  
35.2   / {95.9} / 49.2 \\
ACMM~\cite{ACMM}  & 3200x2130 & 
68.5     / 92.5  / 78.1 & 
72.7    / 88.6 / 79.7 & 
70.4     / 90.7 / 78.9 & 
78.4     / 96.4 / 86.1     & 
83.9     / 96.2 / 89.5 & 
80.9     / 96.3 / 87.7 \\
\hline
CSPN~\cite{cspn}                & 960 x 640 & 
34.9     / 20.0      / 25.1     & 
24.9     / 18.9     / 20.3     & 
30.3     / 19.5     / 22.9     & 
51.9     / 38.2     / 43.4     & 
40.1     / 36.2     / 37.0     &  
46.5     / 37.2     / 40.4 \\
NLSPN~\cite{nlspn}              & 960 x 640 & 
{42.8} / 27.9      / 33.4     & 
\F{35.9} / 29.4     / 32.1     & 
{39.6} / 28.6     / 32.8     & 
{58.0} / 47.3     / {51.9} & 
\F{53.3} / 49.0     / 50.9     &  
{55.8} / 48.1     / {51.4}\\

Ours                            & 960 x 640 & 
\F{47.1} /  \F{32.5} / \F{37.9} & 
{35.8} / \F{30.9} / \F{32.6} & 
\F{41.9} / \F{31.7} / \F{35.5} & 
\F{62.4} / \F{52.8} / \F{56.7} & 
{51.9} / \F{52.3} / \F{51.7} &  
\F{57.6} / \F{52.6} / \F{54.4} \\

\bottomrule
\end{tabular}
}
\vspace{-2mm}
\caption{
    \textbf{Results on ETH3D High-Res Training Set.}
    Our method outperforms other depth completion pipelines under both thresholds.
    Under 2cm, Gipuma~\cite{Red_black_scheme} gets better performance than all depth completion methods, while under 5cm, both ours and NLSPN~\cite{nlspn} perform better than Gipuma.
    State-of-the-art MVS pipeline ACMM~\cite{ACMM} achieves the best performance under both thresholds.
    All depth completion methods are trained on NYUv2. 
    \textbf{Bold} shows the method with highest scores among depth completion networks.
}
\label{tab:eth3d_pc}
\vspace{-0.1in}
\end{table*}

\subsection{Completion from keypoints vs. uniformly random depth samples on NYU v2}
\label{sec:depth_completion_keypoints_nyu}



NYU Depth V2~\cite{nyuv2} consists of paired RGB images and dense depth maps collected from 464 different indoor scenes with a Microsoft Kinect. Following~\cite{sparse_to_dense}, the training sets are generated by sampling evenly from raw video sequences on training splits, and official test split of 654 densely labeled images are used for evaluation. All ground truth depth are obtained by filling missing values using original acquired raw depth. To exclude depth artifacts on borders, we center-crop the original frames to size $448 \times 608$.  We sample 800 random samples and up to 800 SIFT~\cite{SIFT} keypoints for each image.  We use the same evaluation criteria used in ~\cite{sparse_to_dense}, which evaluates the inferred depth Root-Mean-Squared Error~(RMSE), Relative depth error~(REL), and Percent Inlier metrics ($\delta_{\tau}$) where $\tau$ indicates the threshold of the inlier depth ratio.  


See Fig.~\ref{fig:nyuv2_qualitative_figure} for qualitative results. In Table~\ref{tab:nyuv2_whole_table}, we compare performance of CSPN~\cite{cspn}, NLSPN~\cite{nlspn}, and our method on the NYU v2 dataset~\cite{nyuv2} under these two input settings, which we call ``random'' and ``keypoint'' samples for short. 
\begin{itemize}
    \item When trained on random samples, CSPN and NLSPN perform well when tested on random samples ($\delta_{1.02} = 86.3, 88.6$) but perform very poorly when tested on keypoint samples ($\delta_{1.02} = 46.7, 53.0$).
    \item Compared to above, when trained on keypoint samples, CSPN and NLSPN performance reduces when tested on random samples ($\delta_{1.02} = 71.1, 84.8$) and improves moderately when tested on keypoint samples ($\delta_{1.02} = 55.3, 60.3$).
    \item When all methods are trained on keypoint samples, ours performs better than others when tested on random samples ($\delta_{1.02} = 87.8$) or keypoint samples ($\delta_{1.02} = 70.2$).
\end{itemize}

These results indicate that prior methods have difficulty when sparse values are sampled from keypoints, while our method is more effective in this setting.  Further, our method performs as well or better when training on keypoint samples than random samples in either test case, suggesting that keypoint sampling provides a more difficult training regime that leads to a more robust model.

\subsection{Depth completion from sparse SfM points on ETH3D}
\label{sec:depth_completion_keypoints_eth3d}

ETH3D~\cite{eth3d} is a challenging MVS dataset that provides stereo images, sparse point clouds generated by COLMAP~\cite{colmap}, and ground truth depths captured by laser scanners. We use the training set of ETH3D, which contains 13 indoor and outdoor scenes with high-resolution images captured with sparsely sampled views, and evaluate three models trained on NYUv2 with keypoint samples, \emph{without fine-tuning}.  To test, we project the visible points (i.e., reconstructed tracks) from the sparse point cloud into each image to obtain the sparse depth map.  
These sparse depth values contain some outliers due to surface reflections and repeated structures, which makes the depth completion even more challenging. We downsample the images and the corresponding ground truth depth to $960 \times 640$ (from original $6048\times 4032$), due to memory and runtime constraints. 

We fuse the point clouds using the completed depths from each method by checking the absolute depth distance between each reference image with the source views following the fusion strategy of~\cite{Red_black_scheme,Schnberger2016PixelwiseVS}.
We set the depth consistency threshold to 0.1 meters and number of required consistent views to 2. 

We compare the point clouds obtained using the spatial propagation methods to GIPUMA~\cite{Red_black_scheme} and ACMM~\cite{ACMM} MVS algorithms in  Table~\ref{tab:eth3d_pc} using standard metrics from ETH3D. 


\renewcommand{\arraystretch}{1.5}
\begin{figure}[t]
    \setlength{\belowcaptionskip}{-0.3cm}
    \centering
    \begin{tabular}{c@{}c}
        \includegraphics[align=c,width=0.22\textwidth]{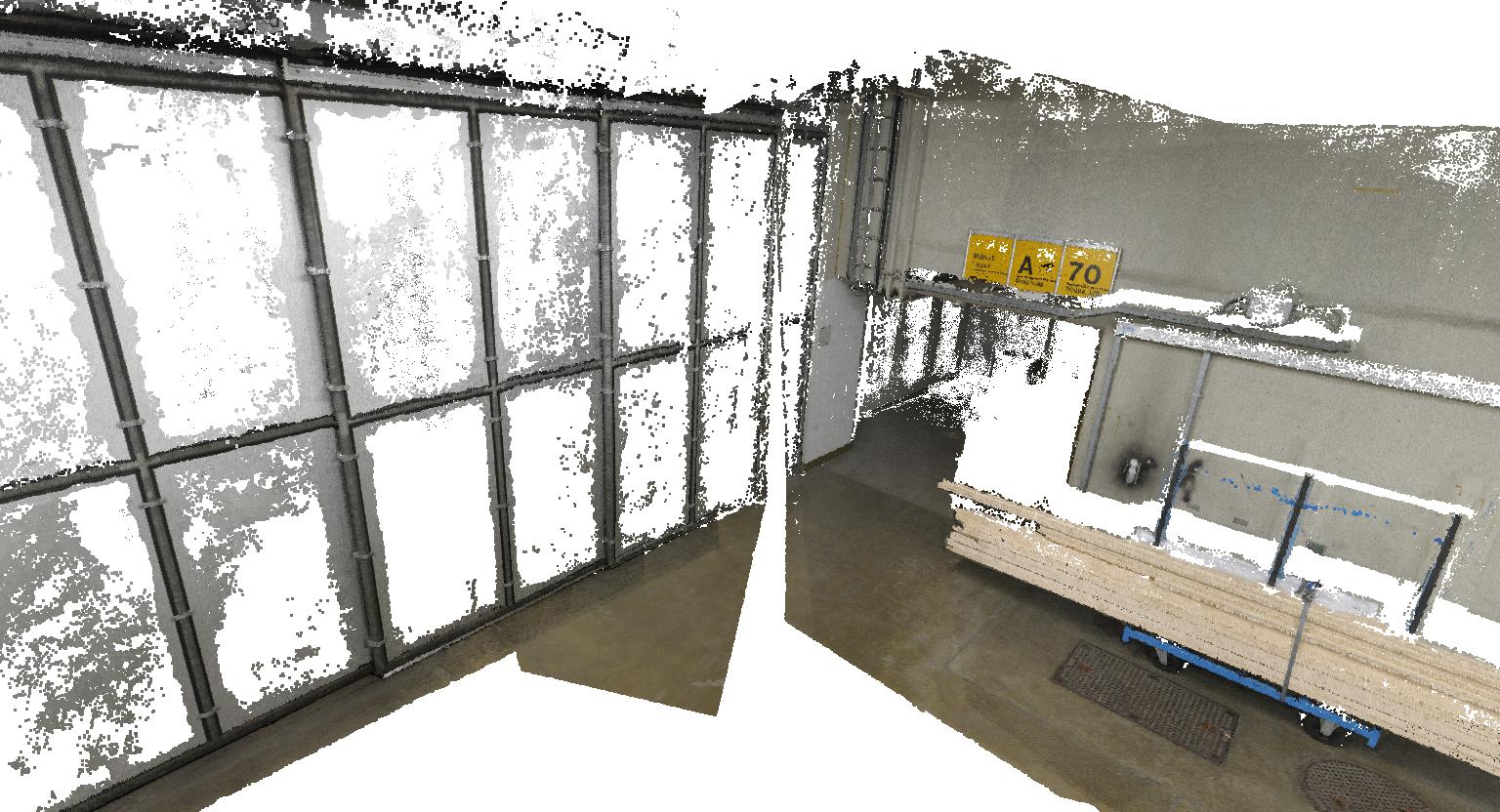} &
        \includegraphics[align=c,width=0.22\textwidth]{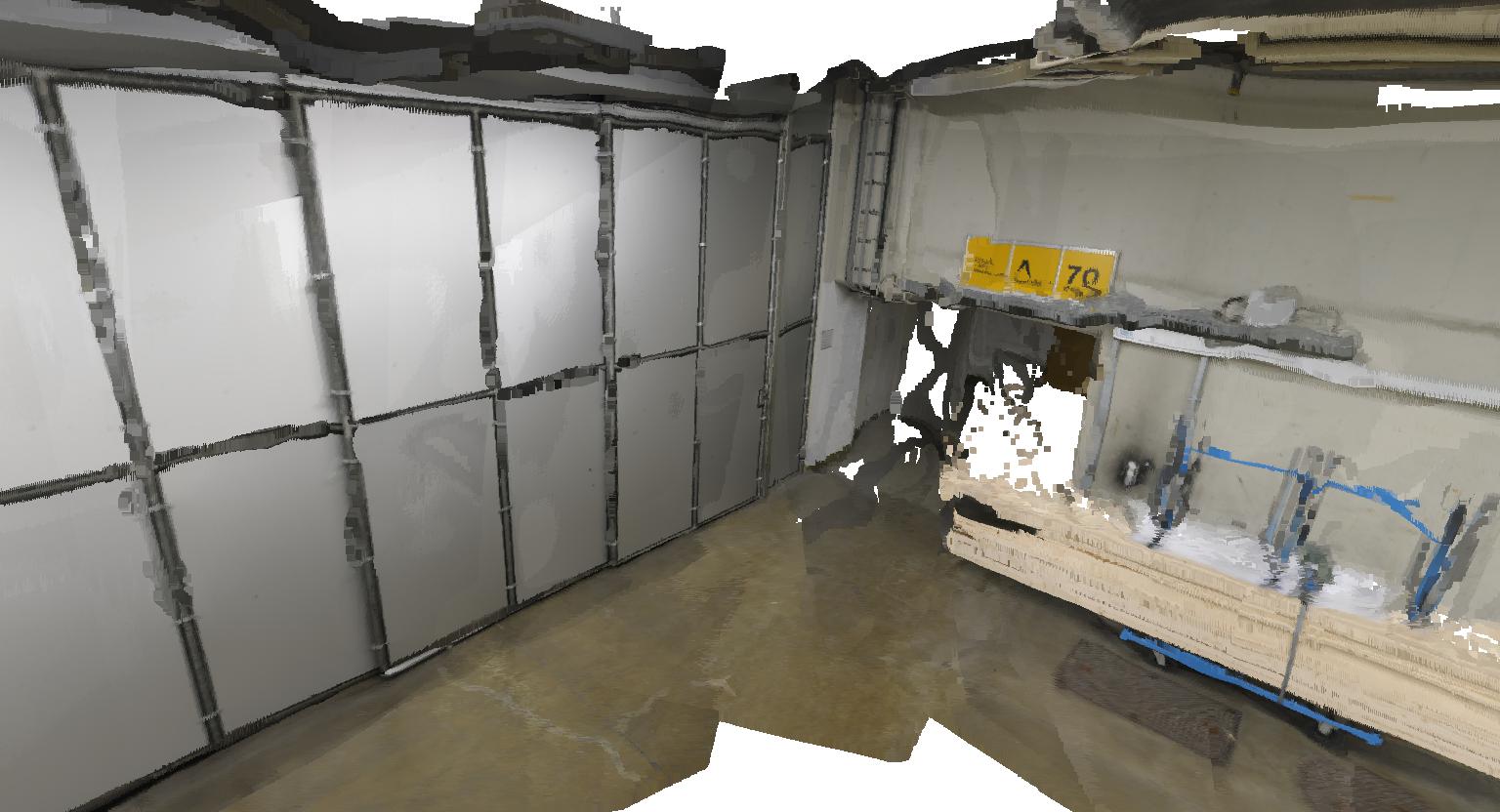} \\
        \includegraphics[align=c,width=0.22\textwidth]{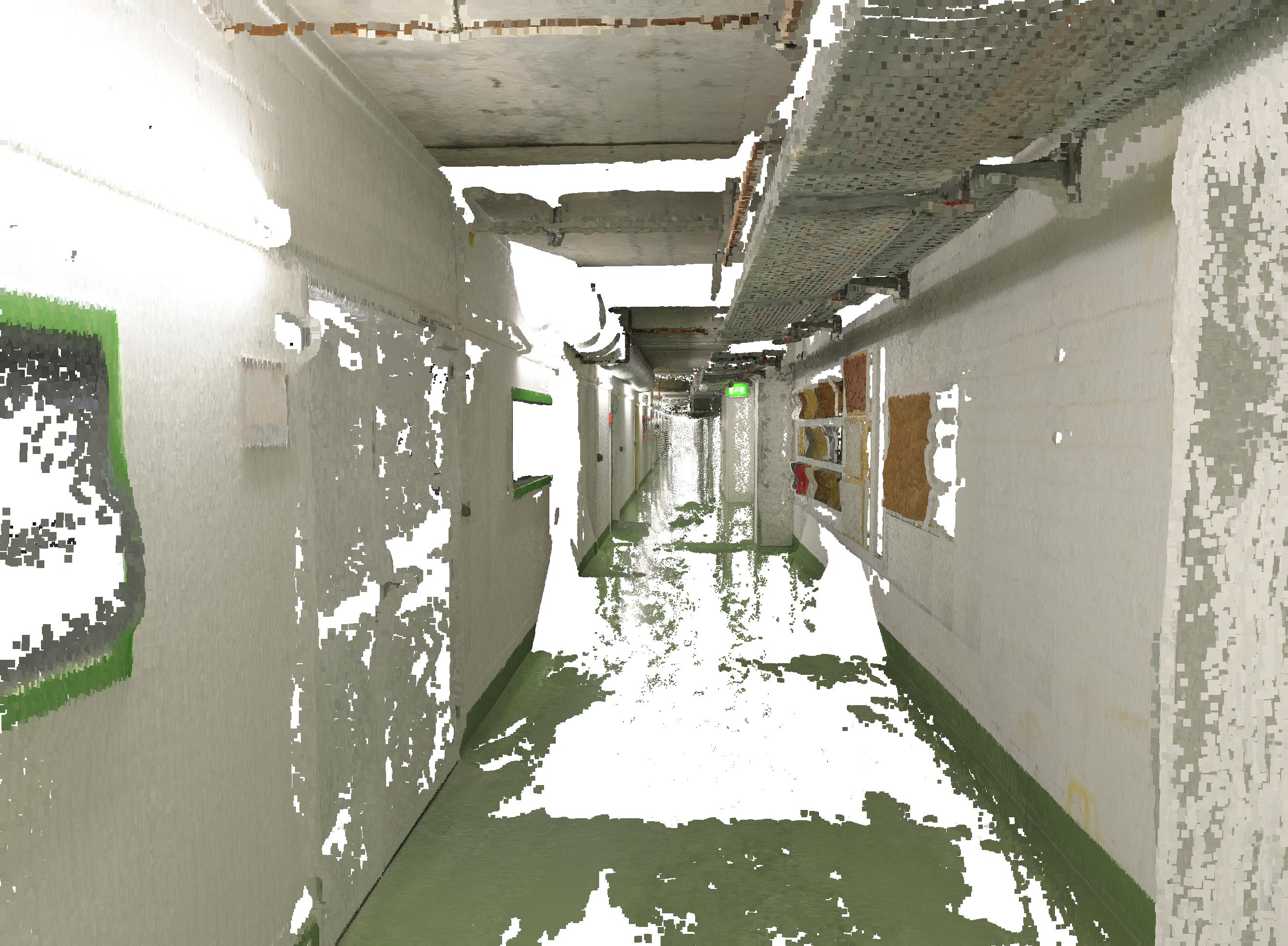} &
        
        \includegraphics[align=c,width=0.22\textwidth]{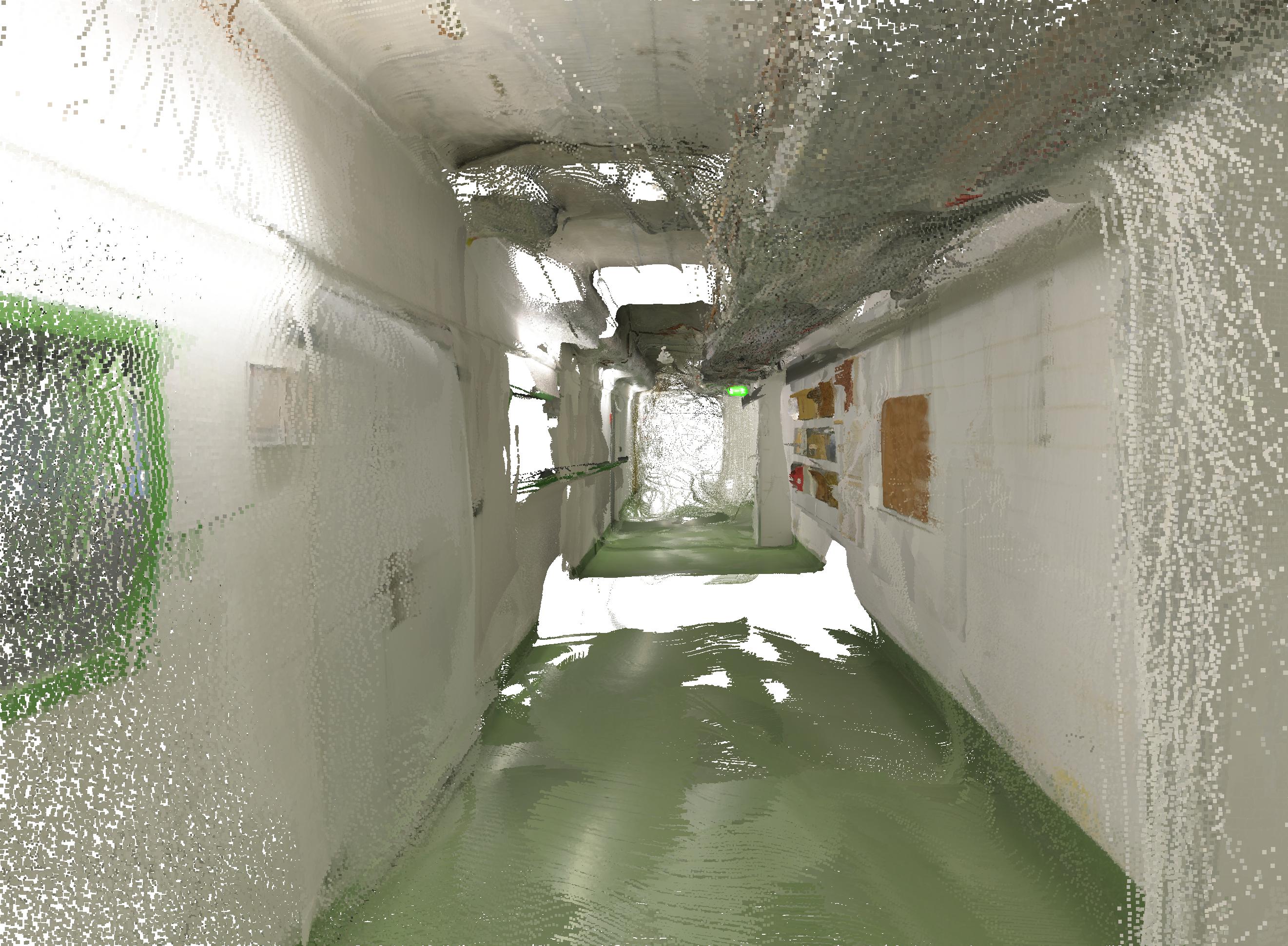}\\
    
        {ACMM~\cite{ACMM}} & {Ours}
    \end{tabular}
    \vspace{-2mm}
    \caption{\textbf{Reconstructed point clouds on ETH3D}. 
    We compare our point clouds fused from single-view completion with the state-of-the-art multi-view stereo ACMM~\cite{ACMM}. Starting with camera poses and sparse points from SfM, our method completes depth in single images and then fuses, while ACMM uses multi-view stereo. Our single-view predictions are less precise than ACMM, but complete texture-less surfaces that are failure points of MVS algorithms.
    }
    \label{fig:eth3d_pc} 
 \end{figure}

\renewcommand{\arraystretch}{1}

\begin{itemize}
    \item The point cloud from our fused single-view image completion achieves similar $F_1$ scores to GIPUMA MVS, with higher completion but lower accuracy.
    \item For combined 2cm and 5cm F1-score, our method (35.5, 54.5) outperforms CSPN (22.9, 40.4) and NLSPN (32.8, 51.4), with a significant improvement on indoor scenes.   
\end{itemize}

Considering Fig.~\ref{fig:eth3d_depth_figure}, we see that our method is more robust to incorrect sparse depths and better completes low-texture regions such as walls and pillars.  Fig.~\ref{fig:eth3d_pc} compares the fused point cloud of our method to state-of-the-art MVS algorithm ACMM~\cite{ACMM}. Though not as accurate as ACMM, our point clouds are more complete in textureless areas. These results show the promise of depth completion as an efficient way to generate depth point clouds from SfM or SLAM when MVS-level precision is not required, e.g. for photo tours and other visualizations.

\begin{table}
\setlength{\belowcaptionskip}{-0.4cm}
\centering
\resizebox{83.5mm}{!}{
\begin{tabular}{c|cc|cccc}
\toprule
Method  & RMSE & REL & $\delta_{1.02}$ & $\delta_{1.05}$ & $\delta_{1.10}$ & $\delta_{1.25}$  \\
\midrule
CSPN                  & 0.220         & 0.043         & 55.3          &	78.3        &	88.9        & 96.1          \\
w/ Dil.               & 0.204         & 0.041         & 55.1          &	77.5        &	89.2        & 96.8          \\
w/ C. to F.           & 0.200         & 0.040         & 57.4          &	78.9        &	89.7        & 96.9          \\
w/ Normals           & 0.188         & 0.034         & 62.6          &	83.2        &	92.0        & 97.4          \\
w/ Conf. mask         & 0.221         & 0.043         & 56.5          &	78.8        &	89.0        & 96.2          \\
w/ Dil.\& w/ C. to F. & 0.159 & 0.031 & 64.8          & 84.4 & 92.9 & 98.1 \\ 
w/ All                & \F{0.147}         & \F{0.026}& \F{70.2} & \F{87.6}          & \F{94.4}          & \F{98.4}          \\ 
\bottomrule
\end{tabular}
}
\vspace{-3mm}
\caption{\textbf{Ablation study on NYUv2 Dataset with Keypoint Samples}. 
    We compare our complete network to CSPN~\cite{cspn} with varying settings. 
    From the top,  ``w/ Dil'' denotes CSPN with the dilated convonlutional kernel, ``w/ C. to F.'' denotes coarse-to-fine, ``w/ Normals'' denotes  surface normal input, ``w/ Conf. mask'' denotes using confidence mask in the spatial propagation, ``w/ All'' denotes using all modules (ours).  
    \textbf{bold} denotes the best result.
}
\label{tab:ablation_study_table}
\end{table}

\subsection{Ablation Studies}
Table~\ref{tab:ablation_study_table} evaluates the contribution of adding each of our proposed components to CSPN~\cite{cspn} on the NYUv2 dataset with the keypoint sampling. 
We find that coarse-to-fine inference on CSPN~\cite{cspn}~(``w/ C. to F.'') improves the performance by a small margin, by 2.1\% on $\delta_{1.02}$ and 0.003 on REL. 
Moreover, we show that while using dilated spatial propagation kernel alone does not improve over CSPN~\cite{cspn}, differing only by 0.002 on REL and 0.016 on RMSE~(``w/ Dil.''), the method achieves far better depth completion results when combined with the coarse-to-fine architecture, improving 9.5\% over the CSPN~\cite{cspn} and 7.4\% over CSPN with coarse-to-fine architecture~(``w/ C. to F.'') on $\delta_{1.02}$.  
We show that adding surface normal input achieves a significant improvement by 7.3\% over CSPN~\cite{cspn} on $\delta_{1.02}$ because it provides geometry guidance for textureless plane areas. 
Confidence masks do not have much effect in NYU v2 experiments because the sparse depth samples are obtained from the same depth map used for ground truth.

{\em Time/memory: } CSPN~\cite{cspn}, NLSPN~\cite{nlspn}, and our SSPN achieve frame rates of 4.8, 7.3, 4.9 FPS, with peak memory usage of 2443MB, 6985MB, 7179MB, respectively when testing on Quadro RTX 6000 on image resolutions of 960 x 640. 
Under the same setup, the pretrained Omnidata surface normal estimation model~\cite{kar20223d} achieves frame rate of 72.9 FPS, with peak memory usage of 4695 MB.

\begin{table}
\centering
\resizebox{83.5mm}{!}{
\begin{tabular}{c|c|ccccc}
\toprule
Networks & REL & $\delta_{1.05}$ & $\delta_{1.10}$ & $\delta_{1.25}$ & $\delta_{1.25^2}$ & $\delta_{1.25^3}$ \\ 
\midrule
CSPN (uniform sampling)$^*$       & 1.461  & 5.1  & 9.0 & 17.0 & 29.5 & 43.3 \\
DCN$^*$   & 0.814  & 15.1  & 28.7 & 57.9 & 82.2 & 91.5  \\
CSPN (keypoint sampling)       & 0.714  & 17.6  & 32.2 & 60.5 & 83.8 & 93.7 \\
Ours   & \F{0.166}  & \F{24.3}  & \F{43.6} & \F{75.3} & \F{92.5} & \F{96.8} \\ 
\bottomrule
\end{tabular}
}


\vspace{-3mm}
\caption{\textbf{Cross-dataset evaluation on Azure Kinect Dataset.}
All networks here are trained on NYUv2~\cite{nyuv2}. 
CSPN (uniform sampling)~\cite{cspn} is trained given 200 uniform sampled points, 
DCN~\cite{sartipi2020deep} is trained on ground truth depths sampled from the correspondences obtained from FAST~\cite{FAST} corners, 
and 
CSPN (keypoint sampling) and ours are trained given 800 keypoints.
$^*$ denotes results reported by \cite{sartipi2020deep}.
Numbers in \textbf{bold} are the best result.}  
\label{tab:azure}
\vspace{-0.1in}
\end{table}

\label{sec:ablation_studies_experiments}

\subsection{Comparisons with existing works}
\label{sec:Previous_evaluation_experiment}
We compare our method to existing depth completion methods~\cite{sartipi2020deep,deep_lidar,depth_normal,cspn,cspn++,nlspn} on the visual SLAM, Lidar, and uniform sampled depth completion tasks, to provide a reference point of our method on the existing benchmarks. Specifically, we evaluate our method on the 
Azure Kinect dataset~\cite{sartipi2020deep}, KITTI depth completion benchmark~\cite{kitti} and resized NYUv2 dataset~\cite{nyuv2} with uniform sampled depth.  

\noindent\textbf{Results on Azure Kinect Dataset}:
Azure Kinect Dataset~\cite{sartipi2020deep} contains 24 datasets in indoor areas. Each dataset comprises color, depth images, IMU measurements, and camera’s poses and triangulated feature positions processed by VI-SLAM. 
Following \cite{sartipi2020deep}, we evaluate our network and CSPN~\cite{cspn}, which are trained using keypoint sampling on NYUv2, on Azure under a resolution of 320 x 240. 
Table ~\ref{tab:azure} presents the cross-dataset performance. 
Our method performs best, and CSPN~\cite{cspn} trained with keypoint achieves significant improvement over uniform sampling and performs better than DCN~\cite{sartipi2020deep}.

\noindent\textbf{Benchmarking on KITTI Depth Completion: } KITTI~\cite{kitti} contains over 93k RGB images and Lidar depth input, and ground truth depth. Following ~\cite{nlspn}, we only use the bottom crop (1216 x 240) of the pairs for training. We use the criteria provided by the dataset~\cite{kitti}, which includes RMSE, Invese RMSE~(IRMSE), Mean-absolute Error~(MAE) and Inverse MAE (IMAE). 
Table~\ref{tab:kitti_test} presents the testing performance acquired from the official website. 
Our method is capable of achieving a similar IMAE result with state of the art.

\begin{table} 
\centering
\resizebox{83.5mm}{!}{
\begin{tabular}{c|cccc}
\toprule
Method  & RMSE & MAE & IRMSE & IMAE  \\
\midrule
CSPN ~\cite{cspn}       & 1019.64           & 279.46            & 2.93              & 1.15              \\
PENet ~\cite{PENet}     & 730.08   & 210.55            & 2.17              & 0.94              \\
CSPN++ ~\cite{cspn++}   & 743.69            & 209.28            & 2.07  & 0.90              \\
NLSPN ~\cite{nlspn}     & 741.68 & 199.59   & 1.99    & 0.84     \\
DYSPN ~\cite{Lin2022DynamicSP}     & \textbf{709.12}& \textbf{192.71}   & \textbf{1.88}     & \textbf{0.82}     \\
Ours    & 849.55            & 200.50& 2.16            & 0.84   \\ 
\bottomrule
\end{tabular}
}
\vspace{-3mm}
\caption{\textbf{Results on KITTI test set}. 
We are close to NLSPN~\cite{nlspn} on MAE and IMAE, which shows that our network is also reliable under the Lidar completion task.
Numbers in \textbf{bold} denote the best results.
} 

\label{tab:kitti_test}
\vspace{-0.05in}
\end{table}
\begin{table}
\centering
\resizebox{83.5mm}{!}{
\begin{tabular}{c|cc|cccc}
\toprule
Method  & RMSE & REL & $\delta_{1.02}$ & $\delta_{1.05}$ & $\delta_{1.10}$ & $\delta_{1.25}$  \\ 
\midrule
DeepLiDAR ~\cite{deep_lidar}        & 0.115         & 0.022         & -             & -                 & -                 & 99.3              \\
DepthNormal ~\cite{depth_normal}    & 0.112         & 0.018         & -             & -                 & -                 & 99.5              \\
ACMNet ~\cite{ACMNet}               & 0.105         & 0.015         & -             & -                 & -                 & 99.4              \\
CSPN ~\cite{cspn}                   & 0.117         & 0.016         & 83.4          & 93.5              & 97.0              & 99.2              \\
NLSPN ~\cite{nlspn}                 & \textbf{0.092}& \textbf{0.012}& \textbf{88.0} & \textbf{95.4}     & \textbf{98.0}     & \textbf{99.6}     \\
Ours                      & 0.112         & 0.015         & 85.8          & 94.5              & 97.4              & 99.3              \\ 
\bottomrule
\end{tabular}
}
\vspace{-3mm}
\caption{\textbf{Results on resized NYUv2 with 500 uniformly sampled depth input.}
All numbers here are reported by original papers, and ``-'' means metrics not reported.
Numbers in \textbf{bold} are the best result}  
\label{tab:nyuv2_uniform_500}
\vspace{-0.18in}
\end{table}

\noindent\textbf{Results on Resized NYUv2 Test Set: } Table~\ref{tab:nyuv2_uniform_500} presents the comparison under a resolution of 304 x 228 with 500 random sampling points, following ~\cite{sparse_to_dense}. Given uniformly sampled points, NLSPN~\cite{nlspn} is the best, though our method is not far behind, differing by 0.003 REL and 2.2\% on $\delta_{1.02}$.

\section{Limitations and Conclusion}
We propose the Sparse Spatial Propagation Network (SSPN) that extends CSPN~\cite{cspn} with a coarse-to-fine architecture, dilated kernel, and surface normal inputs, increasing the range of propagation and leading to much better results for completion from keypoint samples.  Our method also outperforms state-of-the-art NLSPN~\cite{nlspn} when training from keypoint samples.  Our point cloud fusion results indicate that depth completion methods may have applications for quickly creating dense point clouds from SfM or SLAM sparse point clouds.  Further, it would be interesting to explore such methods as a refinement for MVS algorithms to improve completeness in textureless and reflective surfaces.  
The primary limitation of our method, as well as other depth completion methods, is precision, especially in comparison to MVS algorithms. Our multiscale network is also slightly slower than the other convolutional SPNs. 

\noindent
\textbf{Acknowledgements} 
This research is partially supported by NSF IIS 2020227, ONR N00014-21-1-2705, and a gift from Amazon.
We thank Liwen Wu for the discussions and experiments of previous projects that helped shape SSPN.




\clearpage
\bibliographystyle{splncs04}
\bibliography{references}

\end{document}


\title{Supplementary Material: \\  Sparse SPN: Depth Completion from Sparse Keypoints}
\vspace{-5in}
\maketitle





\setcounter{page}{1}


\section{Networks Parameters}
We calculate the number of parameter of each spatial propagation network, as shown in Table~\ref{tab:networks_parameters}. 
Three networks have little difference regarding parameter numbers.
\begin{table}[H]
\centering
\resizebox{70mm}{!}{
\begin{tabular}{c|ccc}
\toprule
Method & CSPN & NLSPN & Ours\\ 
\midrule
\# Params. (M) & 26.85 & 26.88 & 27.04 \\
\bottomrule
\end{tabular}
}
\caption{\textbf{Comparison of the number of network parameters.}
For the convenience of comparison under different setups, we reimplement CSPN~\cite{cspn} and NLSPN~\cite{nlspn}, so the numbers here are slightly different than numbers reported in previous papers.}  
\label{tab:networks_parameters}
\end{table}

\section{Additional Quantitative Results}

\subsection{Quantitative analysis on the long-range propagation kernels.} 
\begin{figure*}
    \centering
    \includegraphics[width=0.68\textwidth]{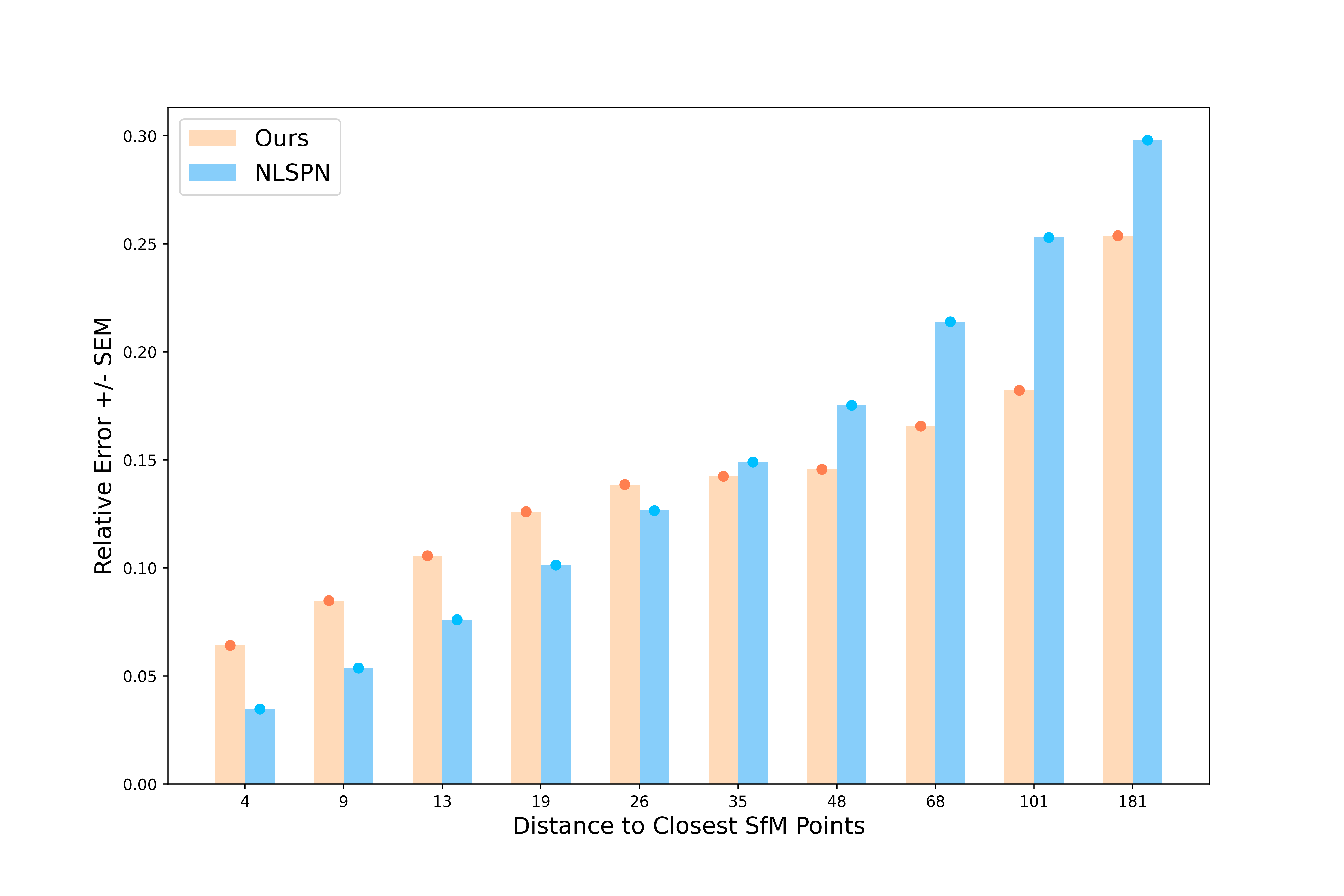}
    \caption{
        \textbf{Distance to closest SfM points vs. relative depth errors on ETH3D.} 
        The x-axis indicates the median distances to the closest SfM points of pixels grouped by percentile. 
        The y-axis indicates the average relative depth errors. 
        The error bar is the standard error of the relative error for each pixel group. 
        All models are trained based on keypoint sampling with NYUv2 Dataset. 
    }
    \label{figure:distance}
\end{figure*}

To validate our contribution on the long-range propagation kernels, we compare our method to baselines on the relative depth error as a function of pixel distance to closest SfM point (Fig.~\ref{figure:distance}). Our method outperforms most when the distance to known depth values is greatest.

\clearpage
\subsection{Depth completion on ETH3D with ground truth values}

To verify the effect of wrong sparse input values, we evaluate performance of each spatial propagation networks on ETH3D when the sparse depth values are replaced with the correct ground truth values.
Table~\ref{tab:eth3d_pc_replaced} presents the point clouds evaluation of CSPN, NLSPN, and our method after replacing the sparse depth values.
All the methods has improvement on the $F_1$ scores at 2cm benchmark by 0.4\%, 1.1\% and 1.3\% for CSPN, NLSPN and Our method respectively, while at the 5cm benchmark, $F_1$ scores for CSPN drops by 0.7\%, NLSPN improves by 0.4\% and ours drops by 0.1\%. 
Note that our method outperforms Gipuma at both 2cm and 5cm benchmarks in this setup.

\begin{table*}[b]
\centering

\resizebox{\textwidth}{!}{
\begin{tabular}{llccccc@{\hskip 0.10in}ccccc@{\hskip 0.10in}ccccc@{\hskip 0.10in}ccccc@{\hskip 0.10in}ccccc@{\hskip 0.10in}ccccc}
\toprule
 &  & \multicolumn{3}{c}{\textbf{2cm: Completeness / Accuracy / F1}} & \multicolumn{3}{c}{\textbf{5cm: Completeness / Accuracy / F1}} \\
\textbf{Method} & \textbf{Resolution}  &  
\multicolumn{1}{c}{\textbf{Indoor}} & 
\multicolumn{1}{c}{\textbf{Outdoor}} & 
\multicolumn{1}{c}{\textbf{Combined}} &
\multicolumn{1}{c}{\textbf{Indoor}} & 
\multicolumn{1}{c}{\textbf{Outdoor}} & 
\multicolumn{1}{c}{\textbf{Combined}}\\
\midrule
Gipuma~\cite{Red_black_scheme}  & 2000x1332 & 
24.6   / {89.3} / {35.8} & 
25.3   / {83.2} / {37.1} & 
24.9   / {86.5} / {36.4} & 
34.0   / {96.2} / 47.1     & 
36.7   / {95.5} / {51.7} &  
35.2   / {95.9} / 49.2 \\
ACMM~\cite{ACMM}  & 3200x2130 & 
68.5     / 92.5  / 78.1 & 
72.7    / 88.6 / 79.7 & 
70.4     / 90.7 / 78.9 & 
78.4     / 96.4 / 86.1     & 
83.9     / 96.2 / 89.5 & 
80.9     / 96.3 / 87.7 \\

\midrule
CSPN~\cite{cspn}                & 960 x 640 & 
34.9     / 20.0      / 25.1     & 
24.9     / 18.9     / 20.3     & 
30.3     / 19.5     / 22.9     & 
51.9     / 38.2     / 43.4     & 
40.1     / 36.2     / 37.0     &  
46.5     / 37.2     / 40.4 \\
NLSPN~\cite{nlspn}              & 960 x 640 & 
{42.8} / 27.9      / 33.4     & 
\F{35.9} / 29.4     / 32.1     & 
{39.6} / 28.6     / 32.8     & 
{58.0} / 47.3     / {51.9} & 
\F{53.3} / 49.0     / 50.9     &  
{55.8} / 48.1     / {51.4}\\

Ours                            & 960 x 640 & 
\F{47.1} /  \F{32.5} / \F{37.9} & 
{35.8} / \F{30.9} / \F{32.6} & 
\F{41.9} / \F{31.7} / \F{35.5} & 
\F{62.4} / \F{52.8} / \F{56.7} & 
{51.9} / \F{52.3} / \F{51.7} &  
\F{57.6} / \F{52.6} / \F{54.4} \\
\midrule
CSPN (GT)~\cite{cspn}                & 960 x 640 & 
35.3   / 20.2 / 25.3     &
24.6     / 19.7     / 20.9     & 
30.3     / 20.0     / 23.3     & 
51.4     / 37.5     / 42.7     & 
38.3     / 35.6     / 36.1     &  
45.3     / 36.6     / 39.7 \\
NLSPN (GT)~\cite{nlspn}              & 960 x 640 & 
{44.8} / 28.6      / {34.5}     & 
{37.4} / 30.4     / 33.2     & 
{41.4} / 29.4     / 33.9     & 
{60.0} / 47.4     / {52.7} & 
\F{53.7} / 48.5     / 50.8     &  
{57.1} / 47.9     / {51.8}\\
Ours (GT)                            & 960 x 640 & 
\F{49.8} /  \F{32.9} / \F{38.9} & 
\F{37.6} / \F{32.8} / \F{34.4} & 
\F{44.2} / \F{32.8} / \F{36.8} & 
\F{63.7} / \F{51.8} / \F{56.6} & 
{51.9} / \F{52.4} / \F{51.6} &  
\F{58.3} / \F{52.0} / \F{54.3} \\

\bottomrule
\end{tabular}
}
\caption{
\textbf{Results on ETH3D High-Res Training Set with ground truth values.}
After replacing the values, our method still outperforms other spatial propagation networks. 
Under 2cm, ours and NLSPN have improvement, and ours outperform Gipuma by small margin. 
Under 5cm, ours has small drop, while NLSPN has small improvement. 
All depth completion methods here are trained on NYUv2. 
GT denotes sparse input with ground truth values.
\textbf{Bold} shows the method with highest scores among depth completion networks with same input.
}
\label{tab:eth3d_pc_replaced}
\end{table*}


\subsection{Robustness of networks per different inputs}
\begin{figure}[b]
    \centering
    \begin{tabular}{c}
        \includegraphics[width=0.70\textwidth]{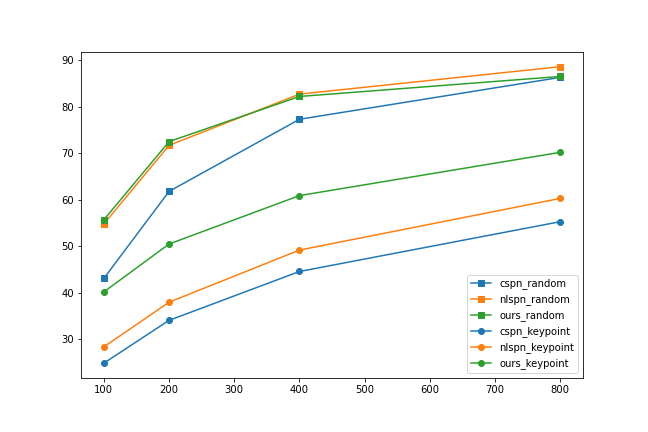} \\
    \end{tabular}
    \caption{\textbf{Performance on NYUv2 given different number of random or keypoint input.} 
    X-axis, Y-axis refer to input numbers and $\delta_{1.02}$.
    Our network is more robust than the other two when given different number of random sampling input. 
    Also, it is more challenging to recover the dense depth when the input is sampled from keypoint than sampled from random, because the latter one has more uniform cover.
    All the networks here are trained given 800 input points sampled from corresponding sampling strategies. 
    }
    \label{fig:nyuv2_diff_num_figure} 
 \end{figure}

We compare each propagation network's robustness to input number in Fig.~\ref{fig:nyuv2_diff_num_figure}. 
The figure shows that keypoint completion is more challenging than random completion, as the accuracy given 800 keypoint inputs is lower than given 200 random inputs.
Also, our method is more robust to the input numbers given both random and keypoint samples.


\clearpage

\subsection{Quantitative analysis of uniform points, SIFT Keypoints, and sparse points obtained from Structure-from-Motion~(SfM).} 
To validate our contribution on keypoint training, we show that SIFT keypoints is a better replacement for reprojected SfM points than uniform sampling.
We validated that the point distributions of SIFT keypoints are more similar to actual sparse depth than uniform samples using precision-recall analysis in Table~\ref{table:distribution}

\begin{table}[h]
    \centering
    \begin{tabular}{ccccccccccccc}
    \toprule
    \multirow{2}{*}{Threshold} & \phantom{a} & \multicolumn{5}{c}{SIFT} &\phantom{a} & \multicolumn{5}{c}{Random}\\
    \cmidrule{3-7} \cmidrule{9-13}
    && $P$ & & $R$ & & $F_1$ & & $P$ & & $R$ & & $F_1$ \\
    \midrule
    3px  && 5.2 & & 4.7 & & 6.9 && 2.8 & & 3.0 & & 2.9\\
    5px  && 25.8 & & 9.1 & & 12.9 && 6.4 & & 7.7 & & 7.0\\
    10px && 47.2 & & 19.6 & & 26.2 && 17.9 & & 26.6 & & 21.2\\
    \bottomrule
    \end{tabular}
    \caption{
        \textbf{Comparison between the distribution of keypoints against the sparse keypoints (obtained via SfM).}
        $P, R, F_1$ denotes precision, recall and $F_1$ scores of the keypoints respectively, where precision indicates percent of target keypoints(SIFT or uniform Random) that contains sparse keypoints within specified threshold and recall indicates the percent of sparse keypoints that contain target keypoints. 
    }
    \label{table:distribution}
\end{table}

\section{Additional Qualitative Results}
We show more qualitative results of point clouds on ETH3D from CSPN, NLSPN, and ours in Fig.~\ref{tab:eth3d_pc_replaced}. 
From the results, CSPN has a poor performance under this setup. 
Our method is generally more complete in texture-less areas and has less noise than NLSPN.
\clearpage
\renewcommand{\arraystretch}{1.5}
\begin{figure*}
    \centering
    \begin{tabular}{c@{}c@{}c}
        \includegraphics[align=c,width=0.3\textwidth]{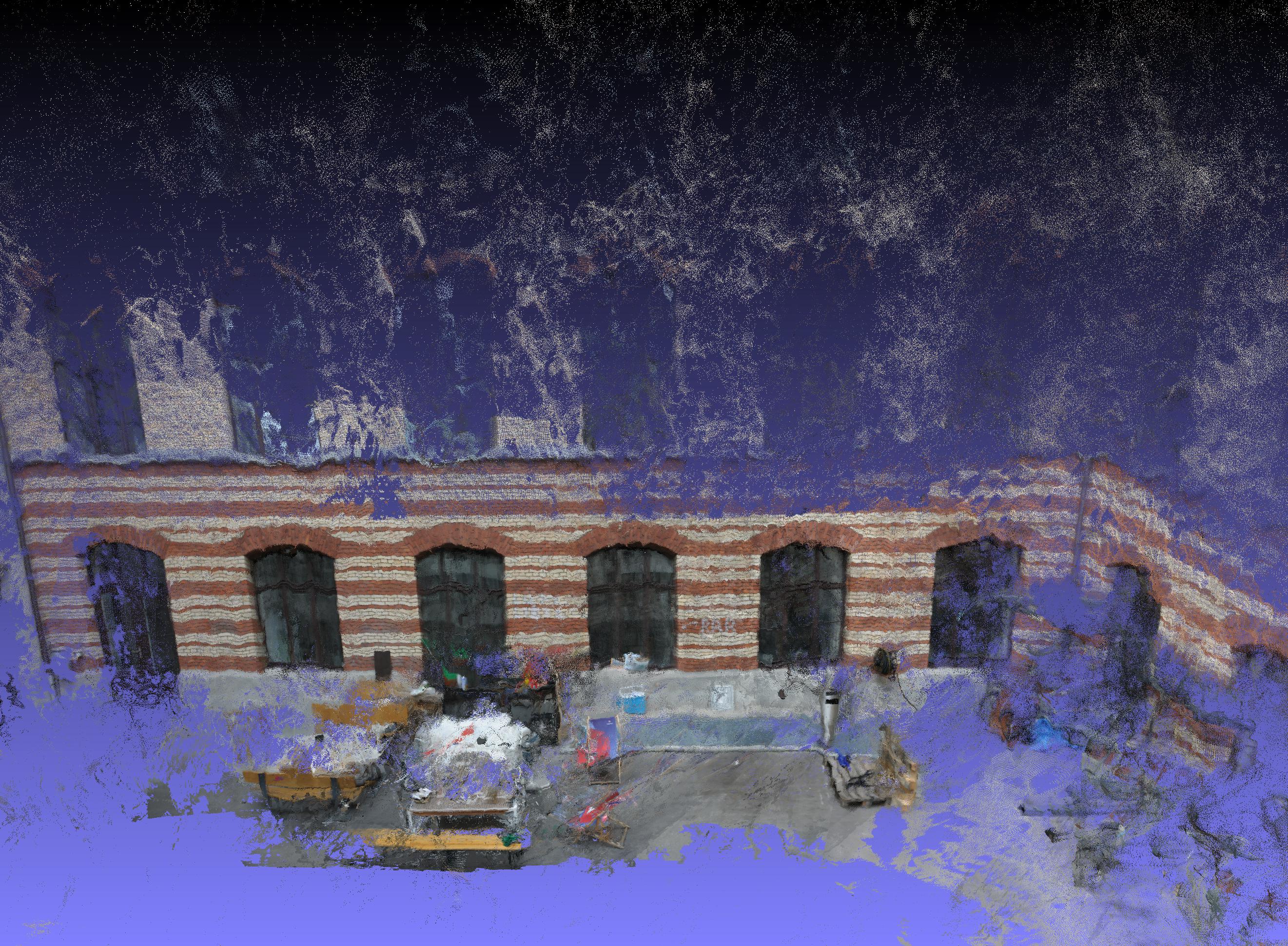} &
        \includegraphics[align=c,width=0.3\textwidth]{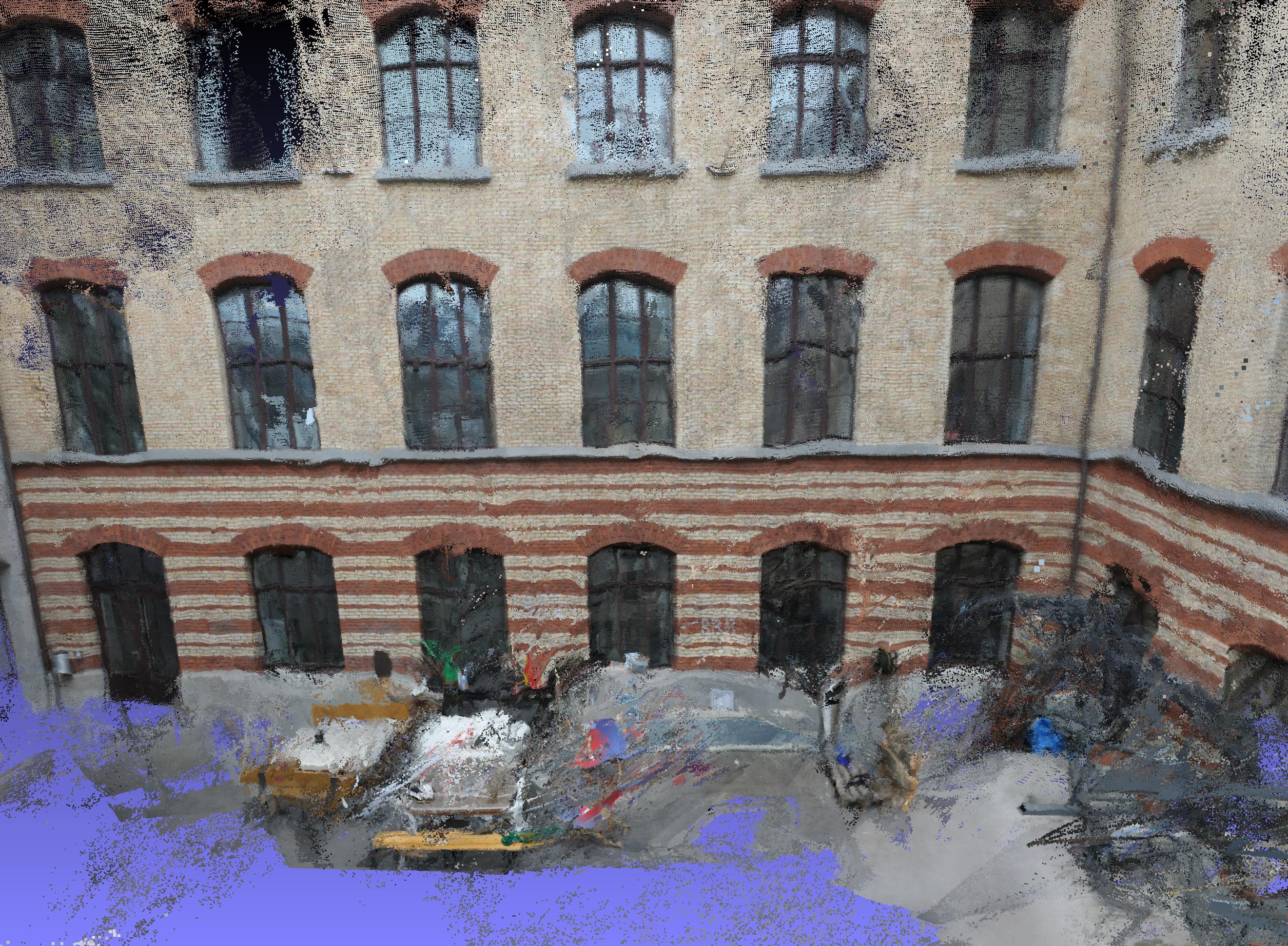} &
        \includegraphics[align=c,width=0.3\textwidth]{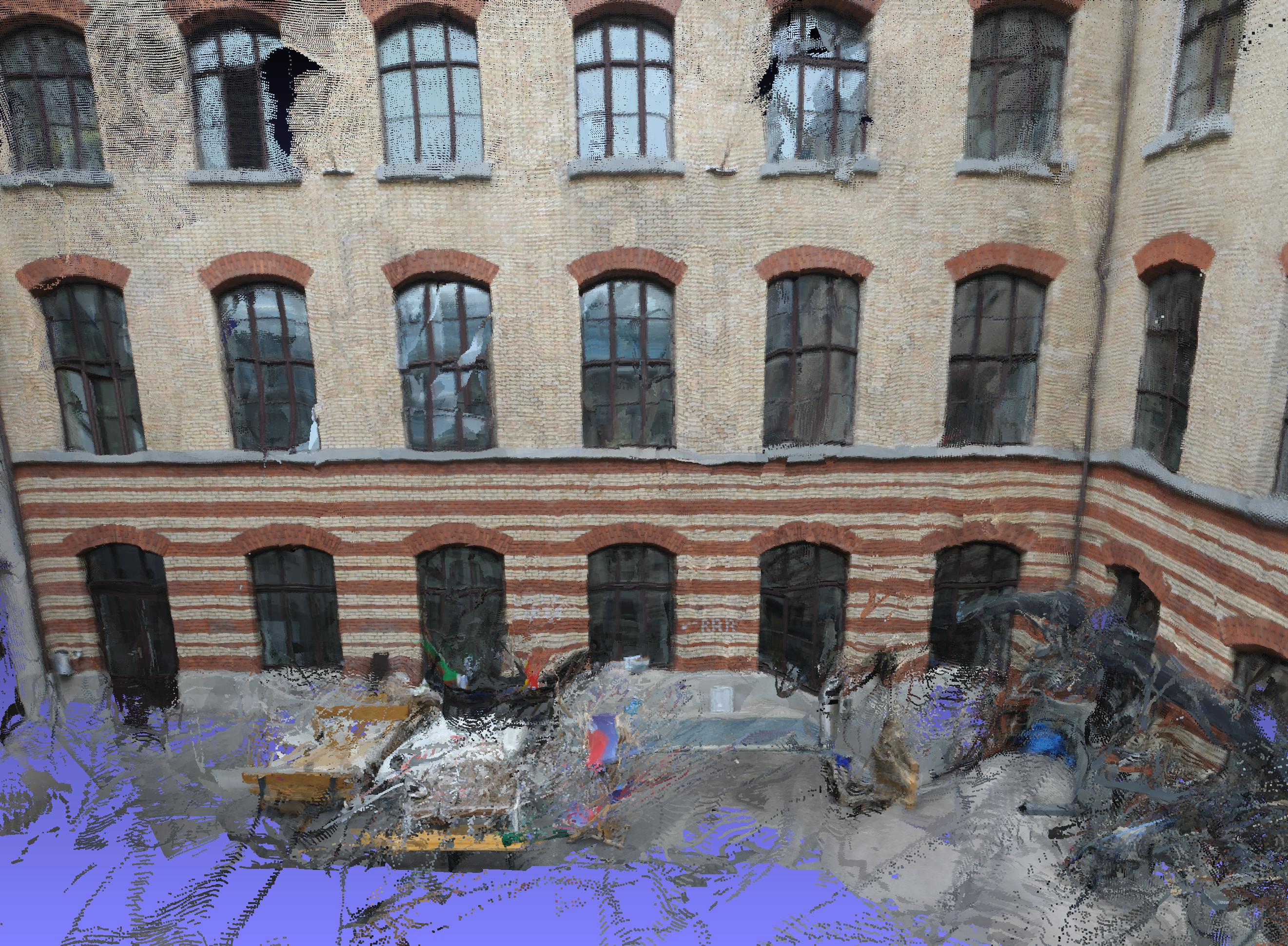} \\
        \includegraphics[align=c,width=0.3\textwidth]{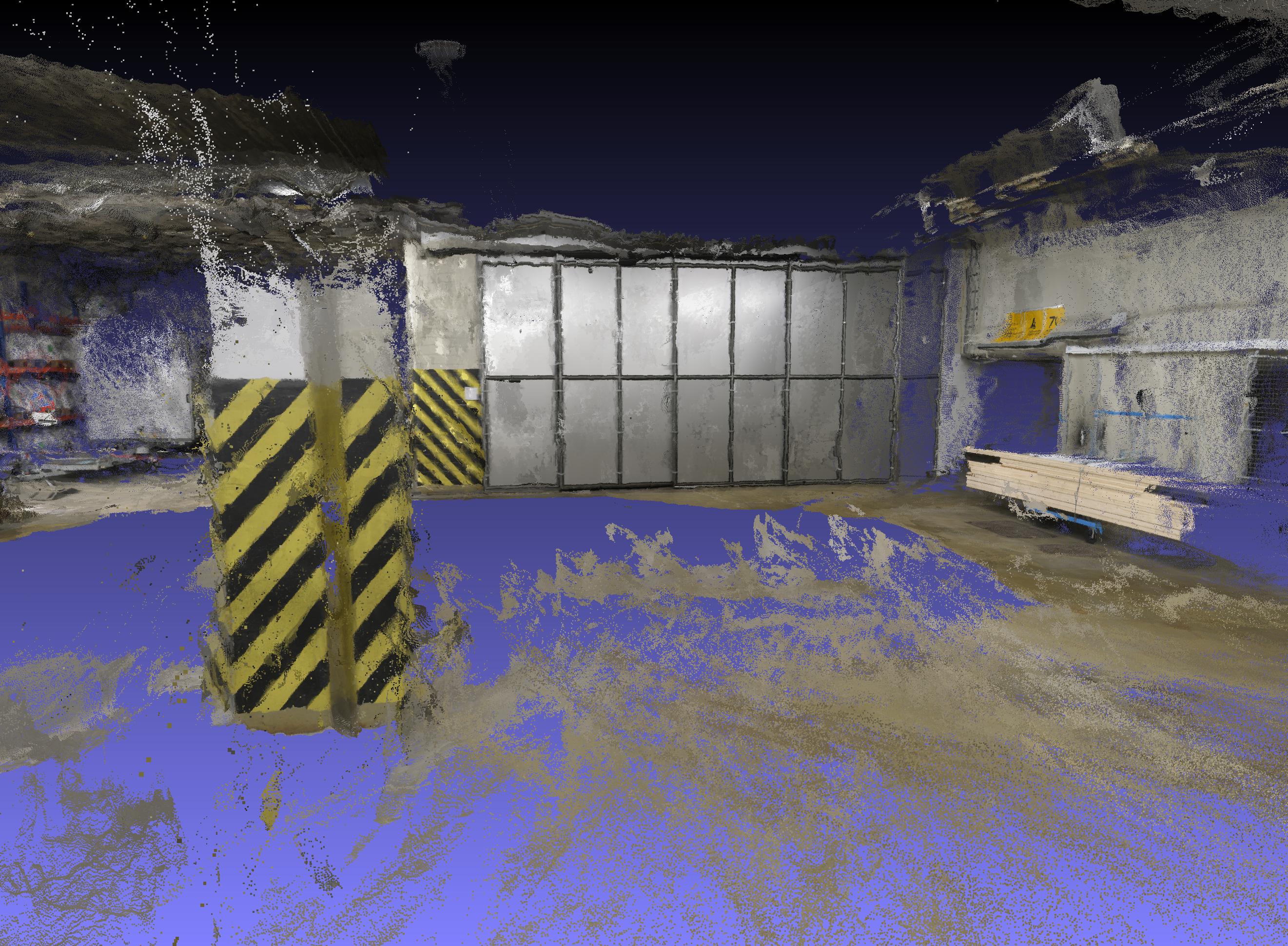} &
        \includegraphics[align=c,width=0.3\textwidth]{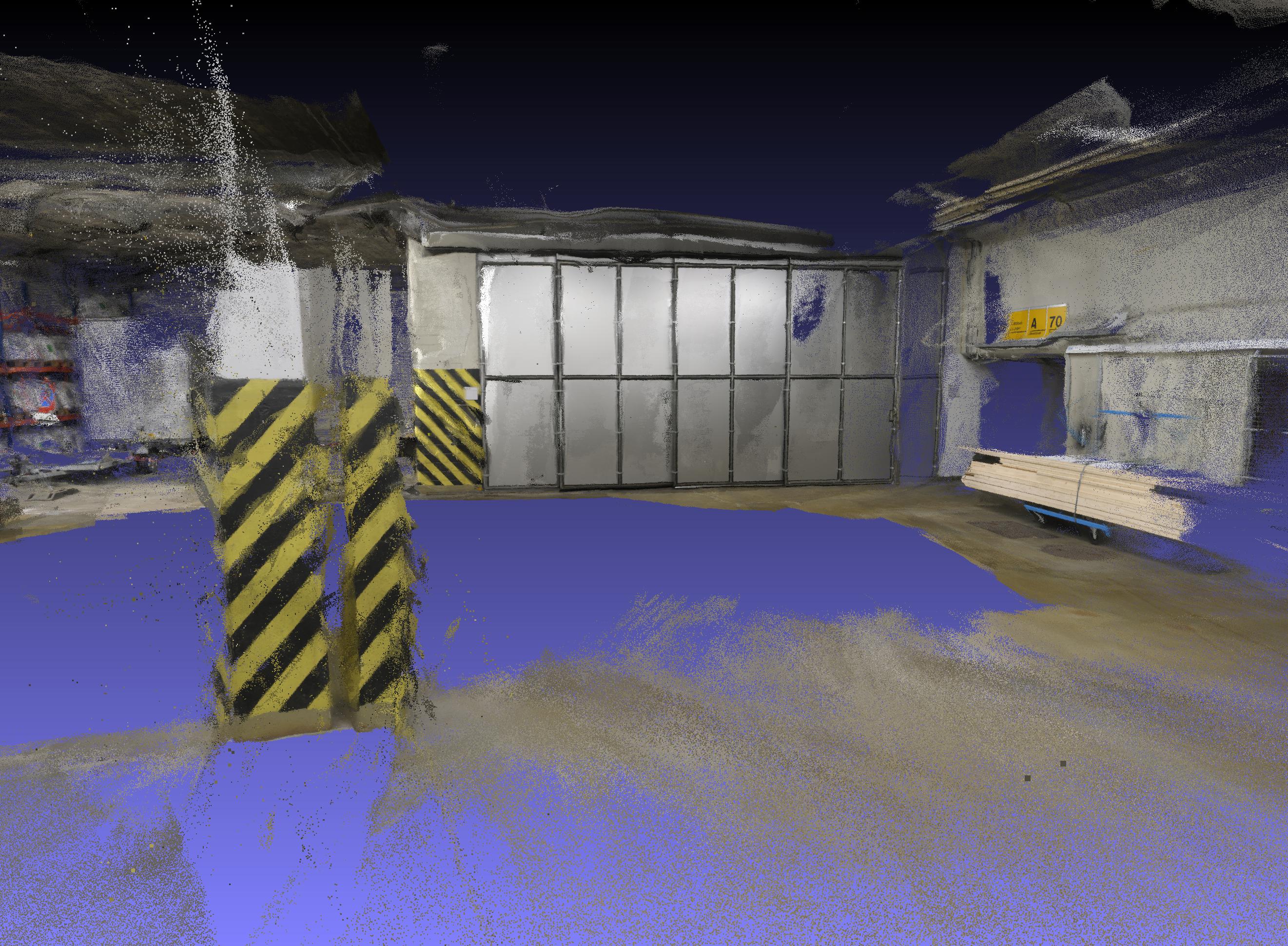} &
        \includegraphics[align=c,width=0.3\textwidth]{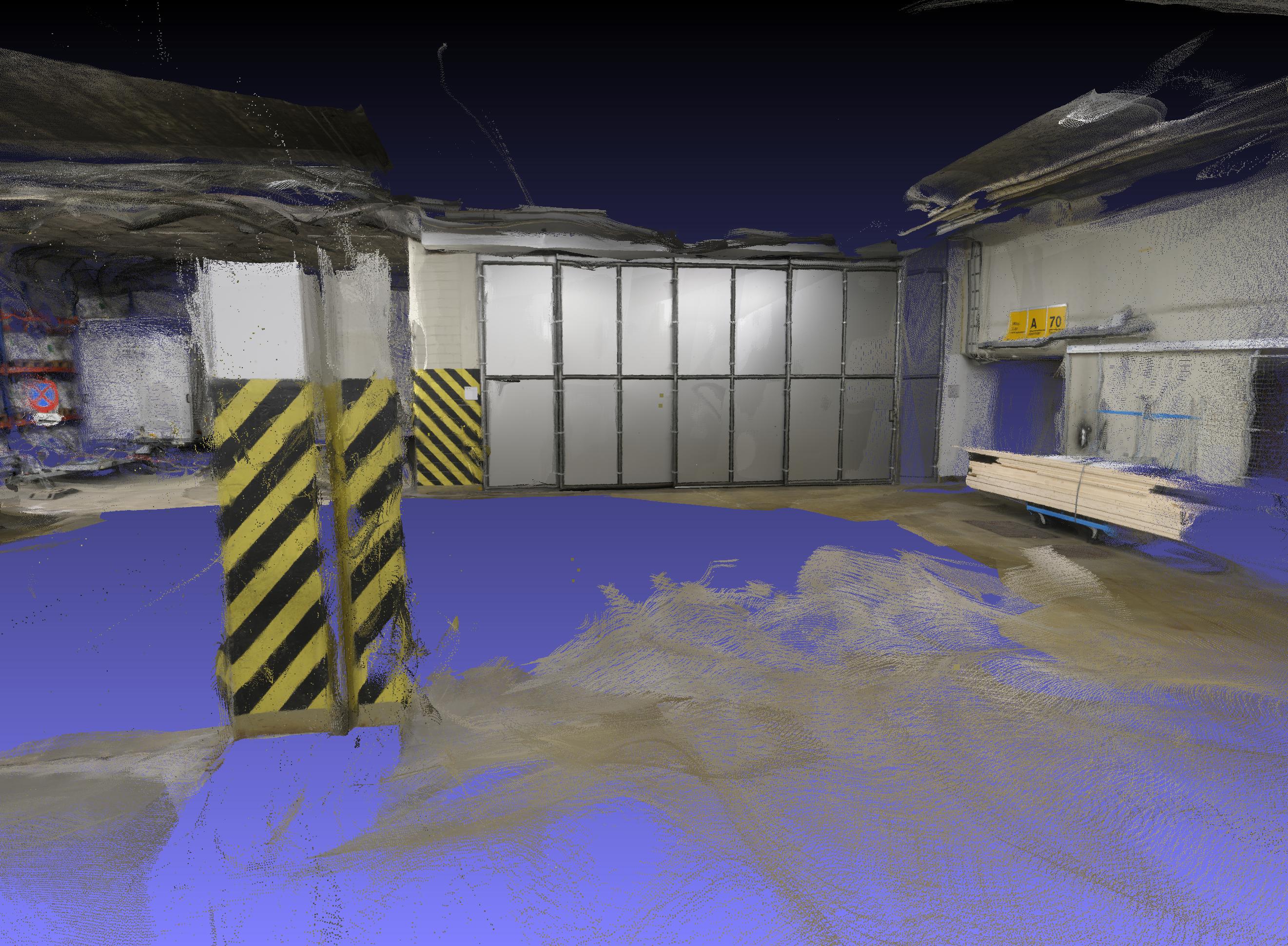} \\
        \includegraphics[align=c,width=0.3\textwidth]{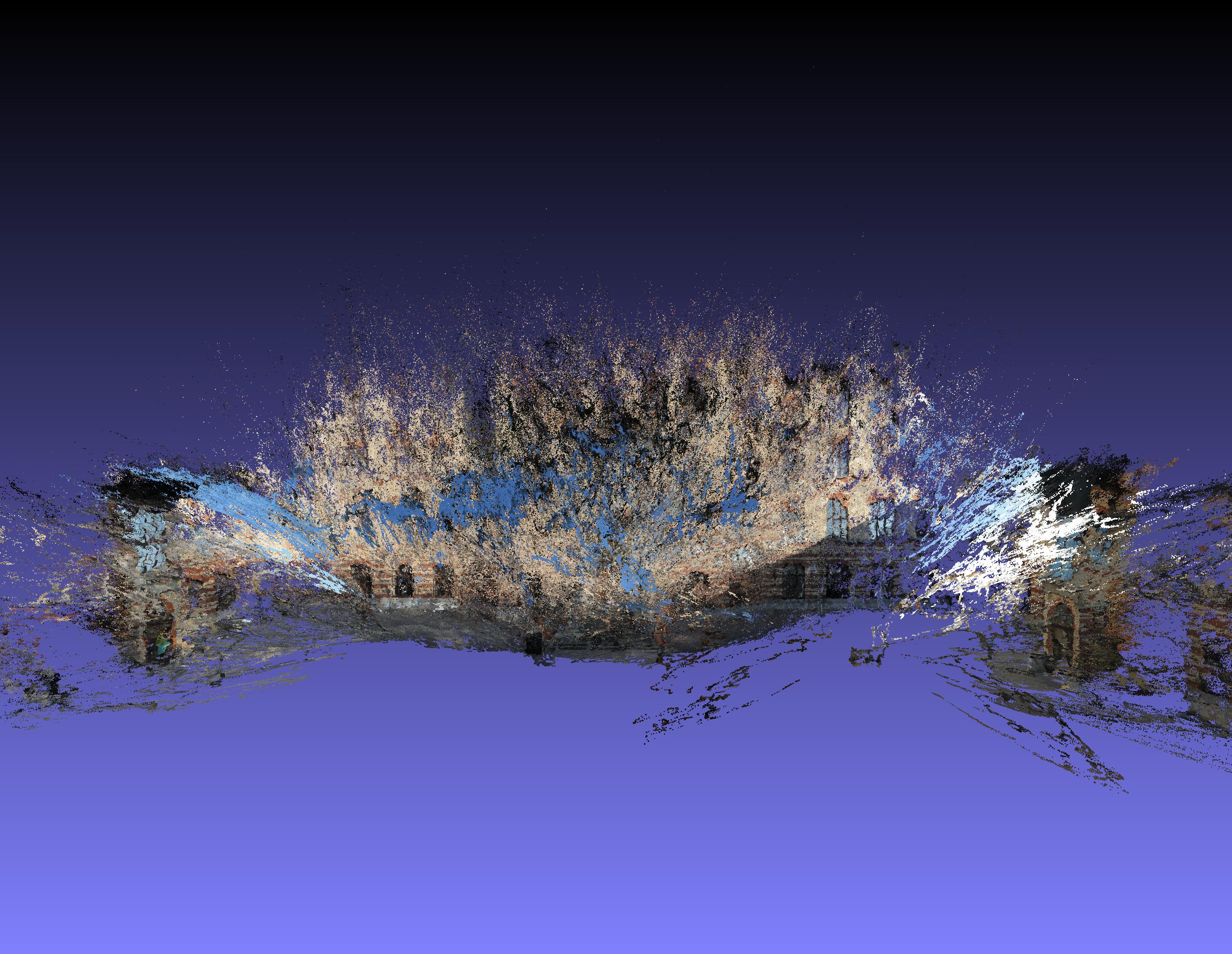} &
        \includegraphics[align=c,width=0.3\textwidth]{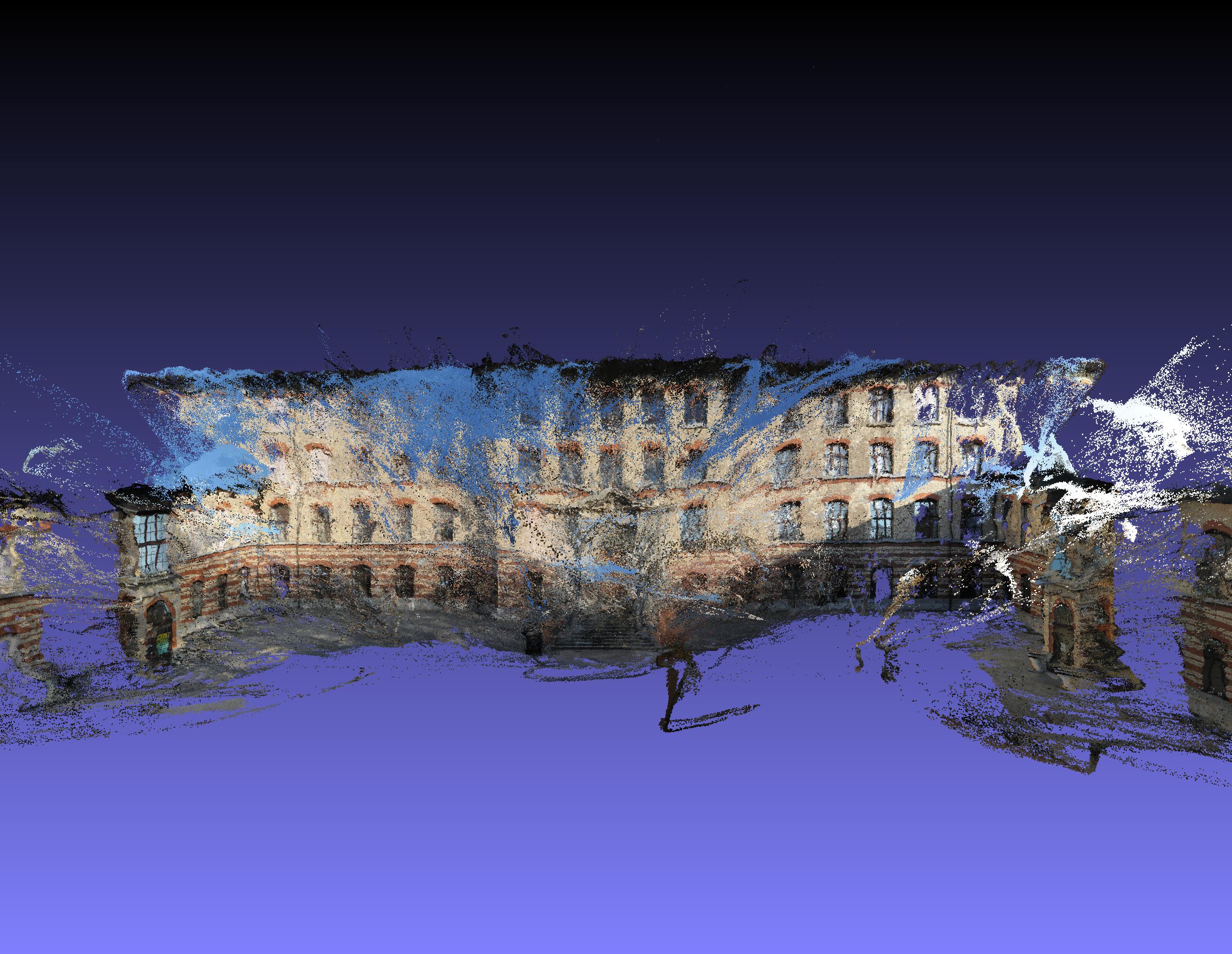} &
        \includegraphics[align=c,width=0.3\textwidth]{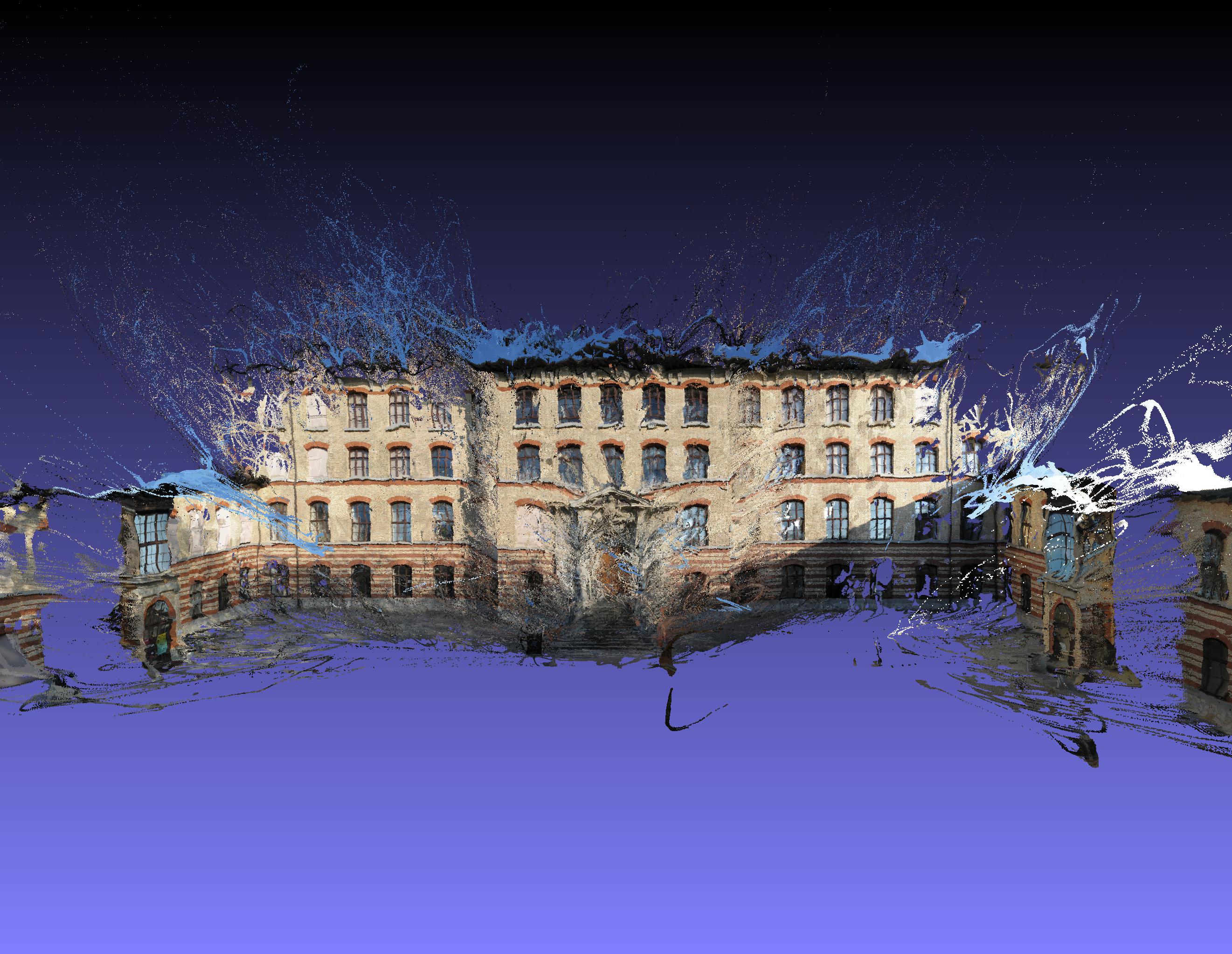} \\
        \includegraphics[align=c,width=0.3\textwidth]{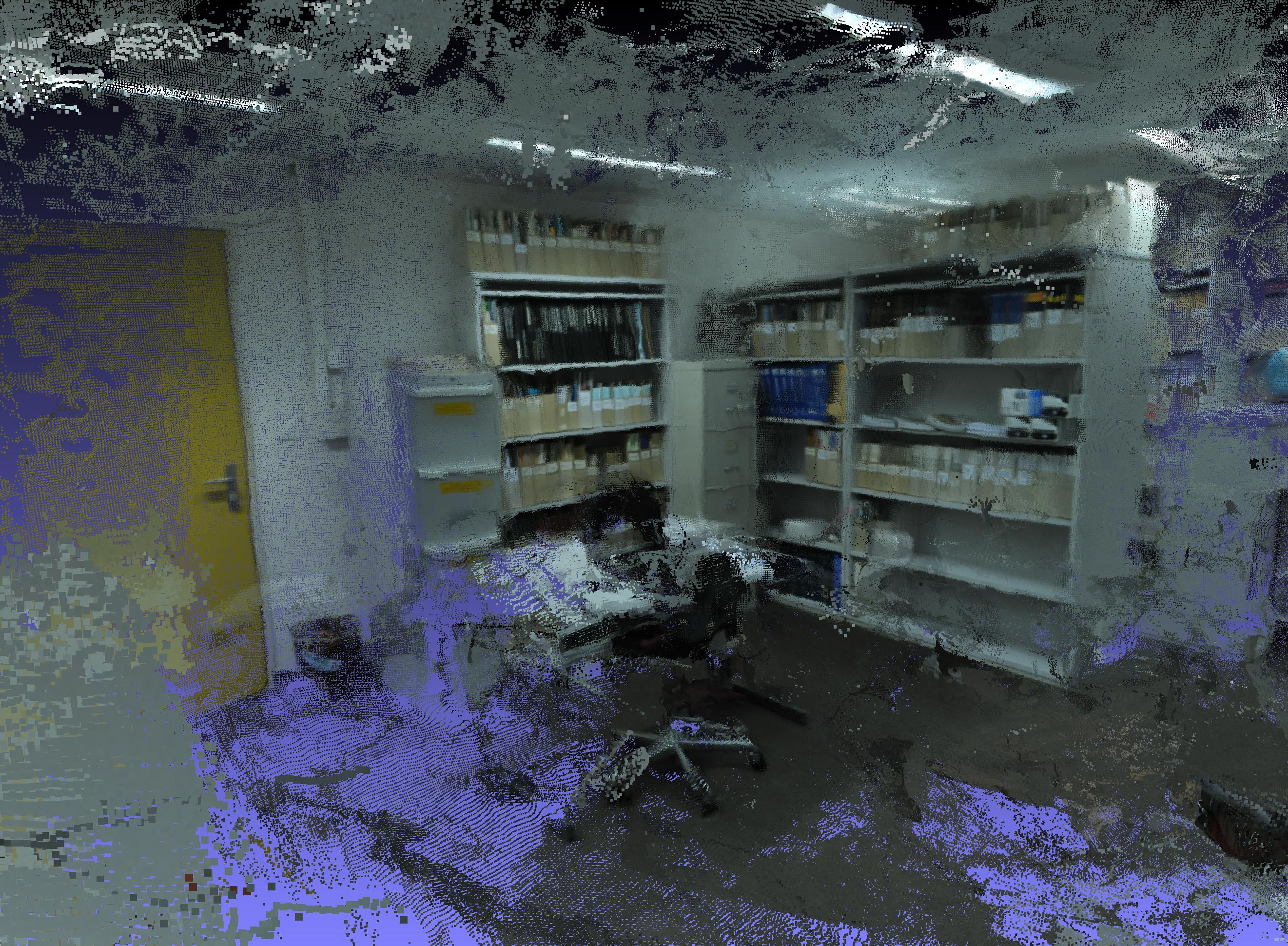} &
        \includegraphics[align=c,width=0.3\textwidth]{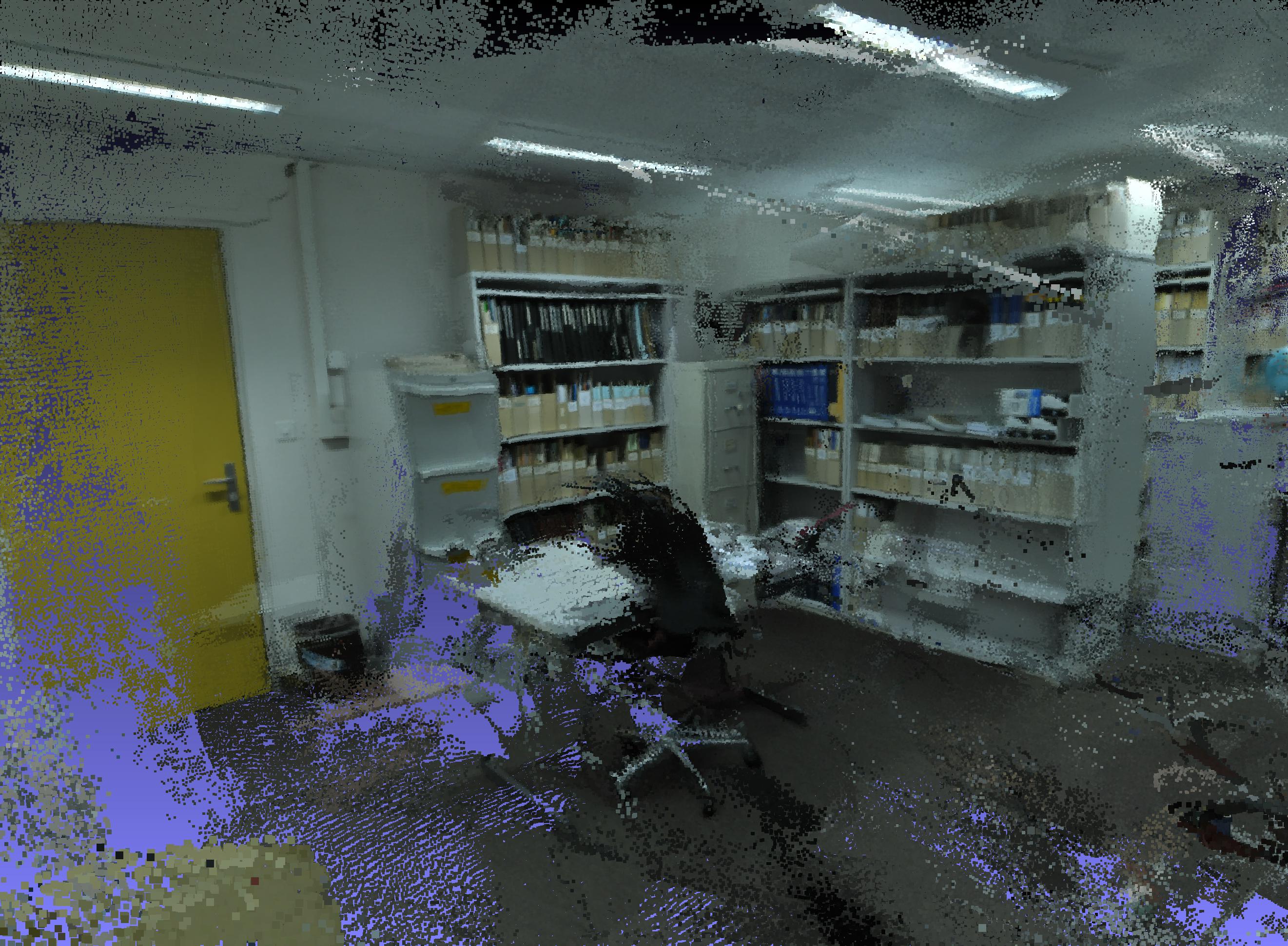} &
        \includegraphics[align=c,width=0.3\textwidth]{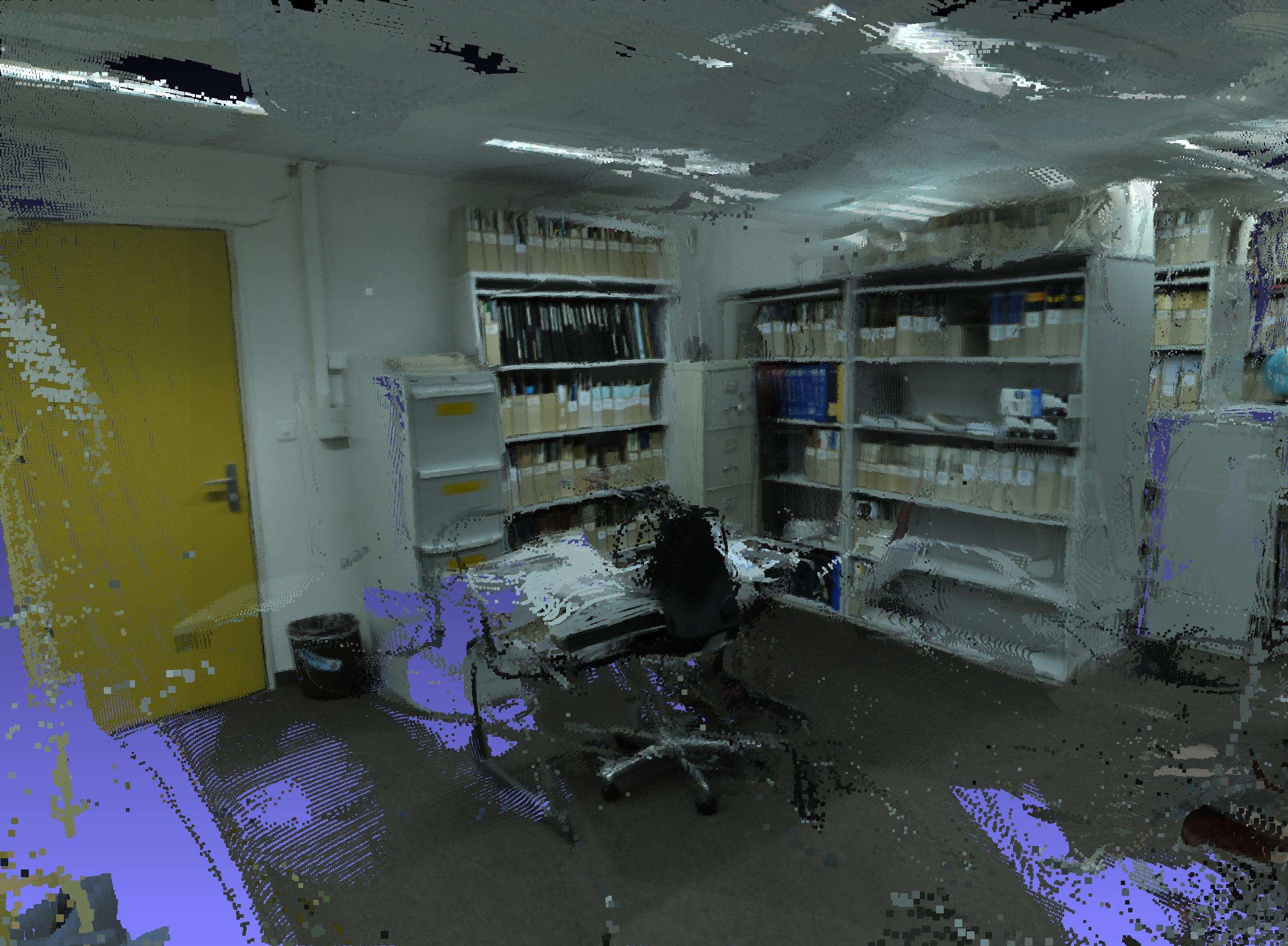} \\
        \includegraphics[align=c,width=0.3\textwidth]{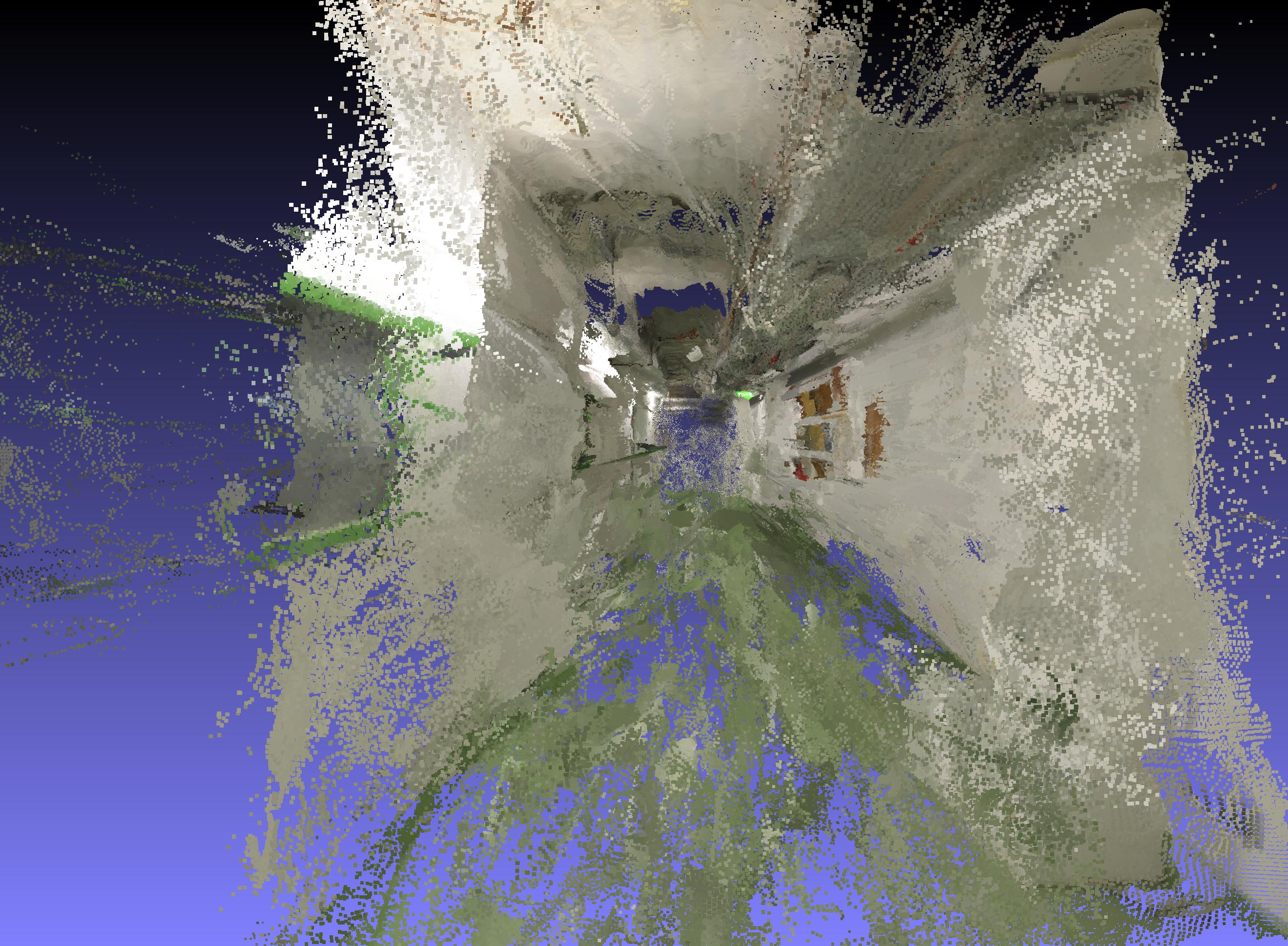} &
        \includegraphics[align=c,width=0.3\textwidth]{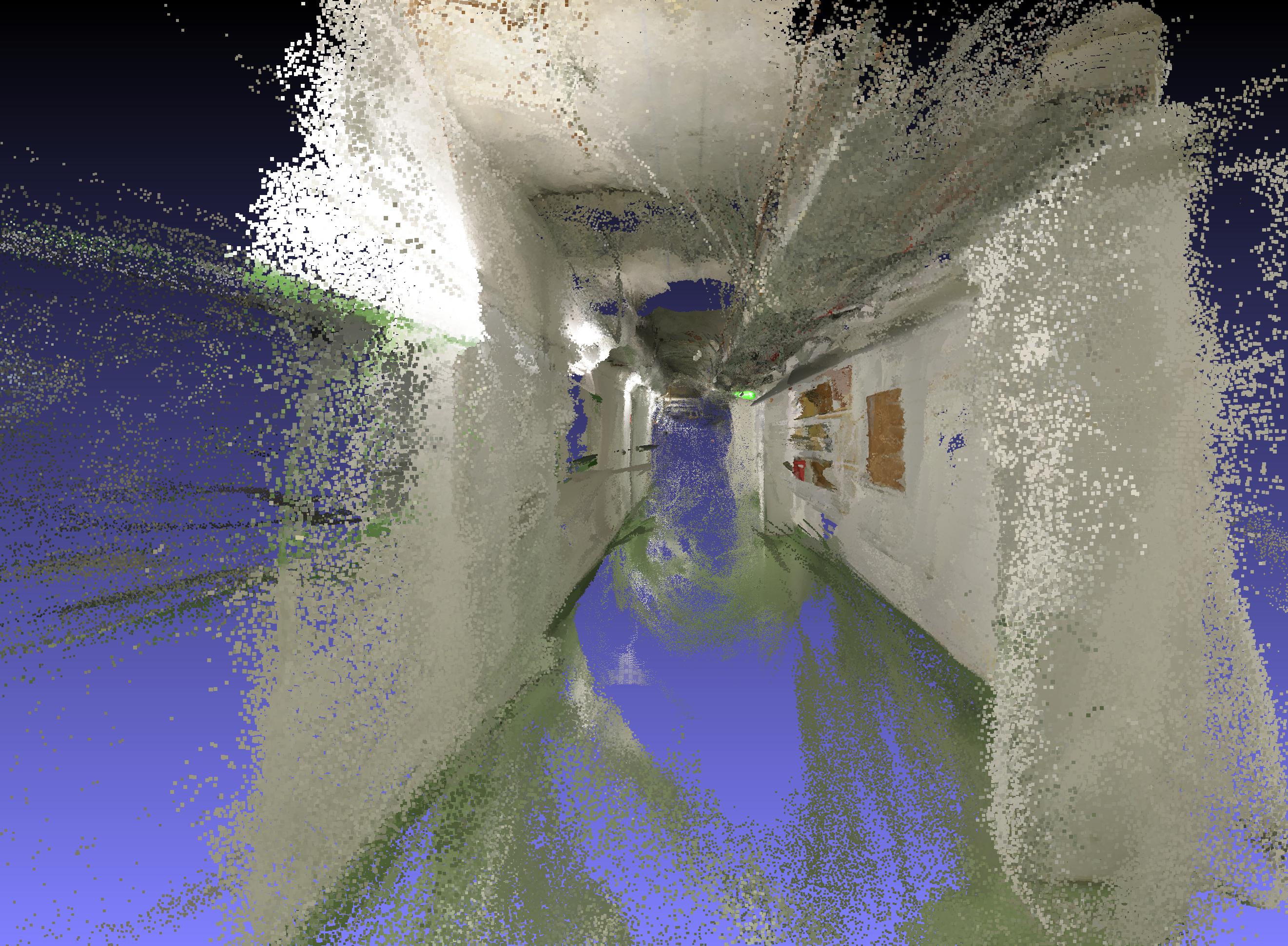} &
        \includegraphics[align=c,width=0.3\textwidth]{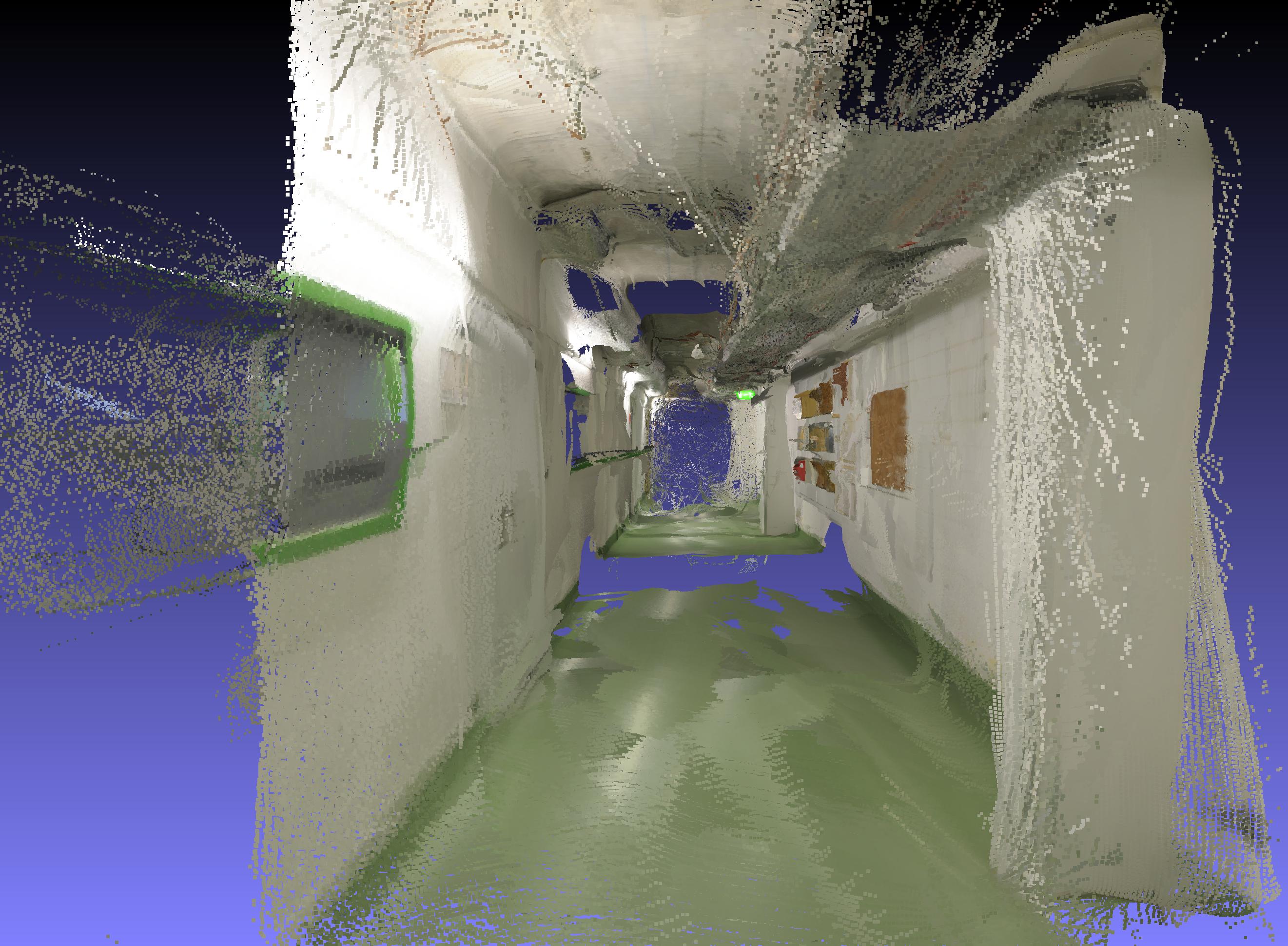} \\
        CSPN & NLSPN & Ours
    \end{tabular}
    \caption{\textbf{Reconstructed Point Clouds on ETH3D}. 
    We show more qualitative results of point clouds on ETH3D of spatial propagation networks. From left to right, CSPN, NLSPN, and ours. Best viewed in zoomed.
    }
    \label{fig:eth3d_pc} 
 \end{figure*}

\clearpage

\clearpage
\bibliographystyle{splncs04}
\bibliography{references}